\documentclass{cambrian}

\usepackage[authoryear,round]{natbib}
\setcitestyle{authoryear,round,citesep={;},aysep={,},yysep={;}}

\definecolor{scholarblue}{rgb}{0.21,0.49,0.74}
\definecolor{darkblue}{rgb}{0, 0, 0.5}
\hypersetup{breaklinks=true,colorlinks=true,citecolor=scholarblue}

\usepackage{thmtools}
\usepackage[normalem]{ulem}
\usepackage{siunitx}
\usepackage{float}
\usepackage{multirow}
\usepackage{makecell}
\usepackage{subfig}
\usepackage{wrapfig}
\usepackage{pdfpages}
\usepackage{soul}
\usepackage[most]{tcolorbox}
\usepackage{fontawesome5}
\usepackage{capt-of}
\usepackage{placeins}
\usepackage{threeparttable}
\usepackage{pgfplots}
\pgfplotsset{compat=1.18}

\usepackage{amsmath,amsfonts,bm}

\def\eqref#1{equation~\ref{#1}}

\def\1{\bm{1}}

\DeclareMathAlphabet{\mathsfit}{\encodingdefault}{\sfdefault}{m}{sl}
\SetMathAlphabet{\mathsfit}{bold}{\encodingdefault}{\sfdefault}{bx}{n}

\newcommand{\cmark}{\ding{51}}
\newcommand{\xmark}{\ding{55}}
\definecolor{deemph}{gray}{0.6}
\definecolor{groupgray}{gray}{0.90}
\definecolor{probeheader}{HTML}{315D7A}
\definecolor{probesubheader}{HTML}{E8F0F5}

\definecolor{proberaegroup}{HTML}{E6DCF2}
\definecolor{probevraerow}{HTML}{F3EEF9}
\definecolor{probevaegroup}{HTML}{F3E2DE}
\definecolor{probevaerow}{HTML}{FAF1EF}
\definecolor{reconlargegroup}{HTML}{DFEAF1}
\definecolor{reconlargerow}{HTML}{F5F8FA}
\definecolor{reconbenchgroup}{HTML}{F2E7D2}
\definecolor{reconbenchrow}{HTML}{FBF7EE}

\definecolor{chartbaseline}{HTML}{8B8397}
\definecolor{chartours}{HTML}{7C5CB2}

\definecolor{findingback}{HTML}{F7F3FC}
\definecolor{findingborder}{HTML}{DCCFEB}
\definecolor{findingaccent}{HTML}{9A7AC7}
\definecolor{findingtag}{HTML}{7253A3}
\definecolor{findingtitle}{HTML}{5C477C}

\newcolumntype{Y}{>{\centering\arraybackslash}X}
\newcolumntype{x}[1]{>{\centering\arraybackslash}p{#1pt}}
\newcolumntype{y}[1]{>{\raggedright\arraybackslash}p{#1pt}}
\newcolumntype{z}[1]{>{\raggedleft\arraybackslash}p{#1pt}}
\newlength\savewidth

\newcommand{\grouprow}[2]{%
  \rowcolor{groupgray}\multicolumn{#1}{@{}l}{\textbf{#2}}\\[-1pt]
}

\newcommand{\finding}[3]{%
  \begin{tcolorbox}[
    enhanced,
    breakable,
    colback=findingback,
    colframe=findingborder,
    boxrule=0.45pt,
    borderline west={2.5pt}{0pt}{findingaccent},
    arc=2.5pt,
    boxsep=0pt,
    left=10pt,
    right=10pt,
    top=7pt,
    bottom=7pt,
    before skip=9pt,
    after skip=9pt
  ]
  \noindent
  \tcbox[
    on line,
    colback=findingtag,
    colframe=findingtag,
    boxrule=0pt,
    arc=1.5pt,
    boxsep=0pt,
    left=4pt,
    right=4pt,
    top=1.5pt,
    bottom=1.5pt
  ]{\color{white}\sffamily\bfseries\scriptsize FINDING~#1}%
  \hspace{0.55em}{\color{findingtitle}\bfseries #2}\par
  \vspace{3pt}
  {\small\color{black!78}#3}
  \end{tcolorbox}%
}

\newcommand{\best}[1]{\textbf{#1}}
\newcommand{\secondbest}[1]{\underline{#1}}
\newcommand{\reldrop}[1]{\,\textcolor{black!55}{\tiny(#1)}}
\newcommand{\na}{--}

\title{\center{V-RAE: Rethinking Video Latent Spaces for Generation}}

\renewcommand\Affilfont{\normalfont\fontsize{11}{15}\selectfont\centering}
\author{%
    \parbox{\textwidth}{%
        \begin{center}
            Minghui Guo\textsuperscript{1} \quad\
            Shengqiong Wu\textsuperscript{2} \quad\
            Hao Fei\textsuperscript{2} \quad\
           \\
            \textsuperscript{1}National University of Singapore \quad \textsuperscript{2}University of Oxford
        \end{center}
    }%
}

\begin{document}

\fancypagestyle{firststyle}{%
    \fancyhead[L]{}\fancyhead[C]{}\fancyhead[R]{}%
    \fancyfoot[L]{%
      \parbox[b]{\textwidth}{%
        \raggedright\footnotesize
        \rule{0.32\textwidth}{0.4pt}\\[3pt]%
        $\dagger$ We sincerely thank Saining Xie for his direct guidance and valuable feedback, which greatly helped shape V-RAE.
      }%
    }%
    \fancyfoot[C]{}\fancyfoot[R]{}%
    \renewcommand{\headrulewidth}{1pt}%
    \renewcommand{\footrulewidth}{0pt}%
}

\let\originalabscontent\abscontent
\renewcommand{\abscontent}{%
  \par\vspace{-26pt}
  \noindent\makebox[\linewidth][c]{%
    \large\textbf{Project Page:}~%
    \href{https://V-RAE.github.io/}{\textcolor{scholarblue}{https://v-rae.github.io/}}%
  }%
  \par\vspace{12pt}
  \originalabscontent
}

\begin{abstract}
Latent video generation relies on autoencoders to define a compact space in which generative models operate.
Although video autoencoder architectures have evolved substantially, their latent spaces are still optimized primarily for pixel-level reconstruction and provide limited high-level semantic organization.
A reconstruction-optimal latent space, however, need not be well suited to generative modeling.
We propose \textbf{V-RAE}, a video representation autoencoder that builds compact generative latents on top of frozen vision foundation model representations.
A lightweight temporal pooling module removes temporal redundancy while preserving semantic structure, and a video decoder reconstructs continuous motion from the compressed features.
We evaluate V-RAE with four representative frozen encoders on video reconstruction, semantic probing, and class-conditional generation.
V-RAE achieves \textbf{2.13 rFVD} on K600, outperforming all evaluated large-scale pretrained video VAEs.
Its latents retain substantially more semantic information than conventional video tokenizer latents.
Under matched generation settings, our best variant achieves gFVD scores of \textbf{117.86} and \textbf{19.16} on UCF101 and K600, respectively, while converging up to \textbf{$6\times$ faster}.
We further show that reconstruction quality alone is insufficient to characterize generative utility and introduce {tFVD}, a temporal-coherence diagnostic that correlates more reliably with downstream generation quality.
Beyond video generation, V-RAE also improves future video prediction on Cityscapes over the Wan 2.2 VAE latent space under matched prediction settings.
Taken together, the experiments show that frozen semantic representations can support video reconstruction, generation, and predictive modeling.
\end{abstract}
\maketitle

\fancyhead[C]{\footerfont V-RAE: Rethinking Video Latent Spaces for Generation}

\par\vspace{0.2em}
\noindent\begin{minipage}{\textwidth}
    \centering
    \captionsetup{font=small,skip=3pt}
    \includegraphics[width=0.98\linewidth]{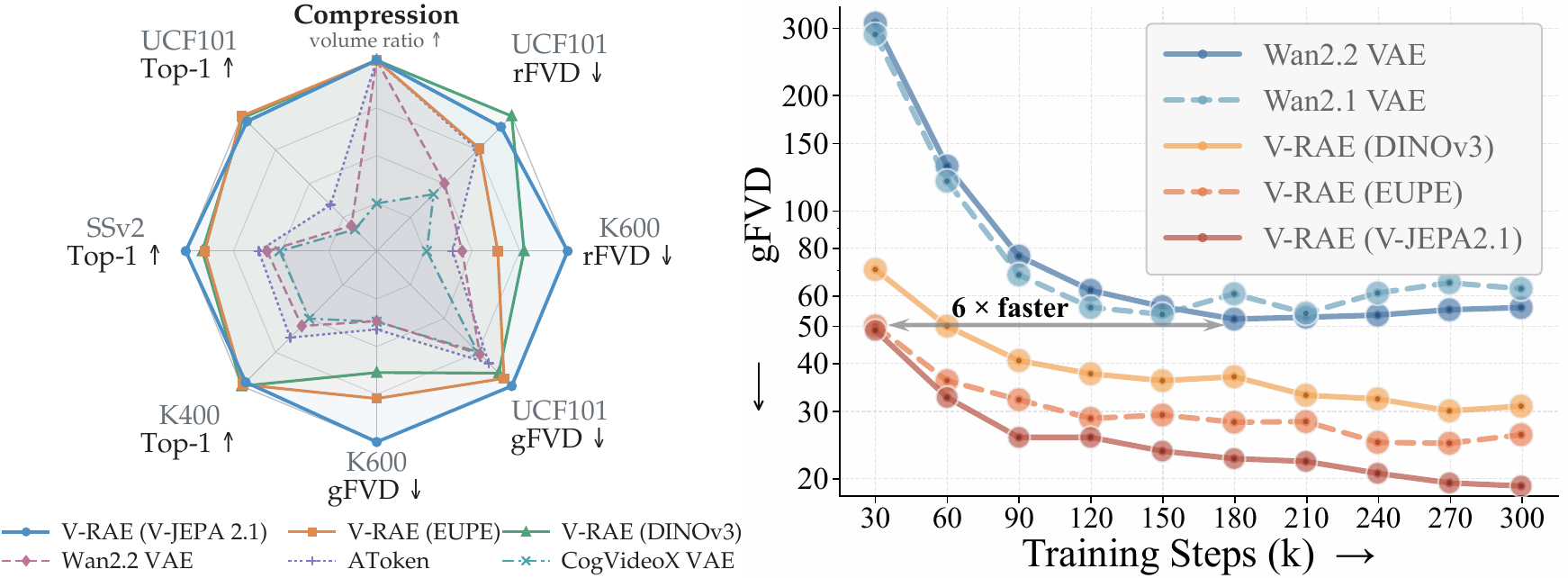}
    \captionof{figure}{\textbf{V-RAE overview.} \textbf{Left:} Normalized reconstruction, generation, compression, and semantic performance across video tokenizers; FVD axes ($\downarrow$) are reversed so outward is better. \textbf{Right:} K600 gFVD convergence; V-RAE converges up to $6\times$ faster than VAE-based latent spaces.}
    \label{fig:hero}
\end{minipage}
\par\vspace{0.4em}

\newpage
{
    \hypersetup{linkcolor=black}
    \tableofcontents
}
\newpage

\section{Introduction}

Modern video generation~\citep{blattmann2023stable} is predominantly performed in a learned latent space produced by a video autoencoder, which compresses raw pixels into compact representations on which a diffusion- or autoregressive-based generative model operates~\citep{peebles2023dit,kondratyuk2023videopoet}.
Such compression is essential for reducing the computational cost of video generation, but it also imposes a strong inductive bias on the generative process, as the autoencoder determines the structure and information content of the latent space, thereby shaping the difficulty of generative modeling and placing an upper bound on the attainable generation quality.
Most existing video autoencoders are optimized primarily for pixel-level reconstruction, including continuous VAEs~\citep{wan2025wan,kong2024hunyuanvideo,yang2025cogvideox}, discrete tokenizers~\citep{yu2024magvitv2,wang2024omnitokenizer}, and query-based designs~\citep{wang2025larp,xiong2026evatok}.
Despite their architectural differences, these methods largely prioritize local textures, colors, and fine-grained appearance details.
While such objectives are effective for reconstructing observed pixels, optimizing reconstruction fidelity alone does not necessarily yield a latent space that is favorable for generative modeling~\citep{yao2025reconstruction,skorokhodov2025improving,xu2026making}.
This mismatch is particularly consequential for video, where the generator must jointly model appearance, motion, and long-range temporal dependencies over long latent sequences.

Motivated by this limitation, some recent studies have begun to incorporate semantic information into latent video generation.
At the autoencoder level, following related advances in image generation, \citet{ma2025unitok} and \citet{wu2025towards} introduce semantic alignment objectives during latent-space learning.
At the generator level, VideoREPA~\citep{zhang2025videorepa} extends representation-alignment objectives such as the REPA loss~\citep{yu2024repae,leng2025repae} to video generation by aligning intermediate diffusion features with pretrained visual representations.
Nevertheless, these approaches primarily use semantic features as auxiliary supervision for the autoencoder or generator, rather than directly defining the generative latent space.
More recently, RAE~\citep{zheng2025rae} and RAEv2~\citep{singh2026raev2} directly adopt the feature space of a frozen image representation encoder as the generative latent space.
The resulting latents are semantically rich, structurally coherent, and well-suited to diffusion modeling, while providing a shared representation for visual understanding and generation.
Whether this paradigm can be effectively extended to video, with a frozen representation encoder supporting both high-fidelity reconstruction and effective video generation, remains largely unexplored.

This extension is non-trivial because video introduces three additional challenges.
\textbf{First}, representation encoders produce temporally dense features.
Image encoders provide no temporal compression, while the native video encoders considered here provide only $2\times$ compression, far below the rate required to keep DiT training and inference affordable.
\textbf{Second}, temporal coherence must be preserved in both compressed latents and decoded motion; otherwise, videos exhibit flicker and jitter, and the latent distribution becomes difficult for diffusion models to learn.
\textbf{Third}, temporal compression must preserve the semantic structure inherited from the encoder, which is the central advantage of using representation features as generative latents.

To address these challenges, we propose \textbf{V-RAE}, a video representation autoencoder that uses frozen visual features as its latent space and adds learnable temporal compression and decoding for video generation.
\textit{On the encoder side}, a frozen VFM encoder integrates with a learnable lightweight temporal pooling module that removes temporal redundancy while preserving semantic structure.
\textit{On the decoder side}, we design a spatiotemporal Transformer decoder equipped with 3D RoPE~\citep{su2021roformer} to reconstruct temporally coherent videos from the compressed latents.
A multi-frame unpatchify layer maps each temporal latent step to multiple consecutive frames, allowing the decoder to accommodate the effective temporal compression ratio of each representation encoder.
Since the representation encoder remains frozen, the resulting latent space stays grounded in its pretrained semantic organization.

We evaluate V-RAE instantiated with four representative frozen encoders, i.e., DINOv3~\citep{simeoni2025dinov3}, SigLIP2~\citep{tschannen2025siglip2}, EUPE~\citep{zhu2026eupe}, and V-JEPA~2.1~\citep{murlabadia2026vjepa2_1}, across reconstruction fidelity, semantic preservation, and generative quality. Figure~\ref{fig:hero} provides a holistic comparison between V-RAE and representative video tokenizers.
For reconstruction, V-RAE with V-JEPA~2.1 attains \textbf{2.13 rFVD on K600}, surpassing all evaluated large-scale pretrained video VAEs. On UCF101, its DINOv3-L variant reaches 6.12 rFVD, ranking second only to Wan2.1 VAE at 6.05.
V-RAE latents also preserve substantially richer semantic information than conventional VAE latents across three probing benchmarks. In particular, V-RAE with DINOv3-L reaches \textbf{89.13\% top-1 accuracy on UCF101}, compared with only 30.83\% for the strongest VAE baseline.
The semantic richness of the learned latent space further translates into improved generation.
Our best-performing variant achieves \textbf{gFVD} scores of \textbf{117.86} on UCF101 and \textbf{19.16} on K600, outperforming all evaluated VAE-based latent spaces while converging up to \textbf{$6\times$} faster under matched training settings.
Across encoder choices, stronger semantic preservation generally coincides with better generation quality, suggesting that the semantic organization inherited from pretrained visual encoders contributes substantially to these gains.
Interestingly, rFVD and gFVD produce markedly different rankings of video autoencoders, with Pearson correlations of 0.200 on UCF101 and 0.473 on K600, indicating that high reconstruction fidelity does not necessarily imply an effective latent space for generation.
We characterize this discrepancy using \textbf{tFVD}, which evaluates temporal coherence by decoding interpolated trajectories in the latent space.
Compared with rFVD, tFVD exhibits substantially stronger Pearson correlations with downstream generation quality (\textbf{0.919 on K600, 0.621 on UCF101}), making it a more reliable and informative diagnostic of the generative suitability of video latent spaces.
Finally, we demonstrate that V-RAE provides an effective latent representation beyond video generation: on future video prediction with Cityscapes, it substantially outperforms the Wan2.2 VAE latent space under the same prediction architecture and training budget.

In summary, our contributions are threefold:
\vspace{-2mm}
\begin{itemize}
\item We propose V-RAE, which builds a generative latent space on frozen vision representation encoders. Temporal pooling and a spatiotemporal Transformer decoder jointly address temporal compression, temporal coherence, and semantic preservation.
\item We introduce tFVD, a generation-oriented diagnostic for video autoencoders, and show that reconstruction fidelity alone does not fully characterize the generative utility of a latent space.
\item We present a systematic study across four encoders with distinct pretraining paradigms, demonstrating reconstruction competitive with large-scale video VAEs, better generation quality across the evaluated VAE latent spaces, and up to $6\times$ faster convergence under matched comparisons.
\end{itemize}

\section{Methodology}
\vspace{-2mm}
\subsection{Video Representation Autoencoder (V-RAE)}

\begin{figure}[!t]
    \centering
    \includegraphics[width=0.99\linewidth]{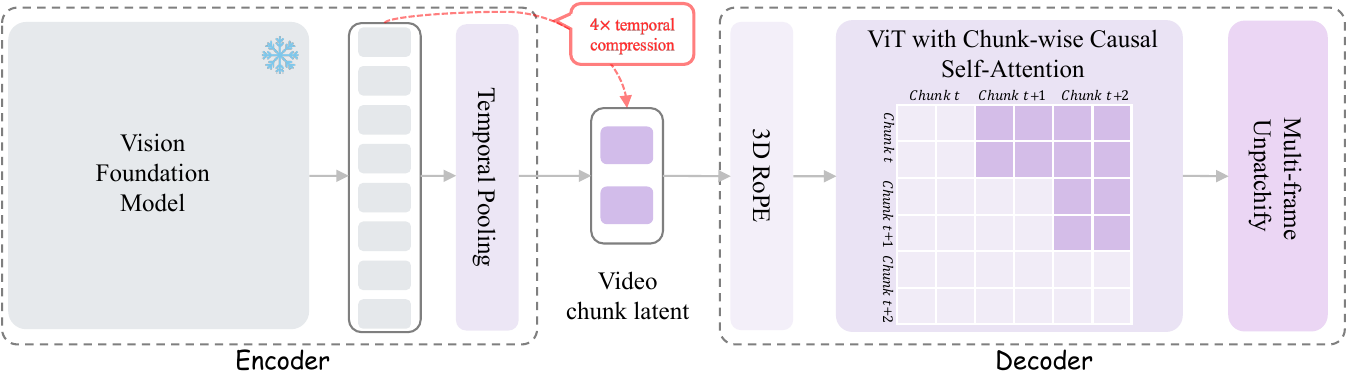}
    \vspace{-2mm}
    \caption{V-RAE architecture. A frozen visual representation encoder produces temporally dense features, which are compressed by a lightweight temporal pooling module. The figure depicts the chunk-wise causal decoder used with image encoders.}
    \label{fig:V-RAE-overview}
    \vspace{-2mm}
\end{figure}

As illustrated in Figure~\ref{fig:V-RAE-overview}, V-RAE couples a frozen visual representation encoder with temporal pooling and a Transformer video decoder.
Temporal pooling compresses visual features into a compact video latent sequence, which is then mapped back to pixels by the decoder.
Only the temporal pooling module and video decoder are learned, while the pretrained representation encoder remains fixed.

\vspace{-1mm}
\paragraph{Visual Representation Encoder.}
V-RAE is designed to operate with frozen visual representation encoders pretrained on either images or videos.
For image-pretrained encoders, including DINOv3~\citep{simeoni2025dinov3}, SigLIP2~\citep{tschannen2025siglip2}, and EUPE~\citep{zhu2026eupe}, we encode each video frame independently and organize the resulting spatial features along the temporal dimension.
These encoders provide strong spatial and semantic representations, but do not natively model cross-frame temporal dependencies.
V-RAE also supports video-native encoders that jointly encode multiple frames into spatiotemporal features.
We instantiate this setting with V-JEPA~2.1-L~\citep{murlabadia2026vjepa2_1}, which builds on predictive video pretraining~\citep{assran2025vjepa2} to produce dense and temporally consistent representations.
This unified interface allows V-RAE to accommodate either frame-wise spatial features or native spatiotemporal features within the same subsequent compression and reconstruction framework.

\paragraph{Temporal Pooling.}
Visual representation encoders often produce temporally dense features with substantial redundancy, resulting in unnecessarily long latent sequences for generative modeling.
We therefore apply a learnable temporal pooling module $\mathcal{P}$ after the frozen encoder to compress adjacent features while preserving their semantic structure.
Given an input video $\bm{X} \in \mathbb{R}^{T \times 3 \times H \times W}$, the encoder produces $\bm{F}=\mathcal{E}(\bm{X})\in\mathbb{R}^{T_E\times C\times h\times w}$, where $T_E$ is the feature sequence length and $r_E=T/T_E$ is the encoder's native temporal compression ratio.
In our experiments, $r_E=1$ for DINOv3, SigLIP2, and EUPE, and $r_E=2$ for V-JEPA~2.1.
The pooler introduces an additional compression ratio $r_P$ by dividing $\bm{F}$ into non-overlapping temporal windows of length $r_P$ and mapping each window to a single latent feature map.
We use temporal attention pooling throughout our main experiments; an architectural illustration and comparisons with alternative pooling designs are provided in Appendix~\ref{app:pooling-designs}.
The resulting latent representation is
\begin{equation}
    \bm{Z} = \mathcal{P}(\bm{F})=[\bm{z}_0,\ldots,\bm{z}_{T_Z-1}] \in \mathbb{R}^{T_Z \times C\times h\times w},\; T_Z=T_E/r_P\;.
\end{equation}
The overall temporal compression ratio is therefore
\begin{equation}
r_{\mathrm{all}} = r_E \cdot r_P = \frac{T}{T_Z}\;.
\end{equation}
After flattening the spatial grid, each latent feature map is represented as
\begin{equation}
\bm{z}_t = \{\bm{z}_{t,p}\}_{p=1}^{hw}\;, \qquad \bm{z}_{t,p}\in\mathbb{R}^{C}\;,
\end{equation}
where $p$ indexes a spatial position.
Similarly, we denote the encoder token at temporal position $s$ and spatial position $p$ by
$\bm{f}_{s,p} \in \mathbb{R}^{C}$.
For temporal attention pooling, the output token $\bm{z}_{t,p}$ is computed from the encoder tokens within the corresponding temporal window,
\begin{equation}
\left\{
\bm{f}_{r_{P} \cdot t+i,p}
\right\}_{i=0}^{r_P-1}\;.
\end{equation}
For notational simplicity, we omit the fixed output index $t$ and spatial position $p$, and write
$\bm{f}_i=\bm{f}_{r_{P} \cdot t+i,p}$\;.
For the $m$-th attention head, the key and value vectors are obtained as
\begin{equation}
\bm{k}_i^{(m)} = \bm{W}_K^{(m)} \cdot \bm{f_i}\;,
\qquad
\bm{v_i}^{(m)} = \bm{W_V}^{(m)} \cdot \bm{f}_i\;.
\end{equation}
The attention weights and pooled features are then computed as
\begin{equation}
\alpha_i^{(m)} = \text{Softmax}_{i}
\left(
\frac{
(\bm{q}^{(m)})^\top \cdot \bm{k_i}^{(m)}
}{
\sqrt{d}
}
+
\beta_i^{(m)}
\right)\;,
\qquad
\bm{u}^{(m)} = \sum_{i=0}^{r_P-1}\alpha_i^{(m)} \cdot \bm{v_i}^{(m)}\;,
\end{equation}
where $\bm{q}^{(m)}\in\mathbb{R}^{d}$ is a 1D learnable query vector shared across temporal windows and spatial positions, $\beta_i^{(m)}$ is a learnable bias associated with the local temporal offset $i$, and $d=C/M$ is the dimension of each of the $M$ attention heads. The pooled output token is
\begin{equation}
\bm{z}_{t,p} =\text{Norm}(\bm{W}_O \cdot \text{Concat}([\bm{u}^{(m)}]_{m=1}^{M}))\;,
\end{equation}
where $\bm{W}_O$ denotes the output projection.
The same operation is applied independently at every spatial position, reducing the temporal resolution without introducing additional mixing across the spatial grid.
We initialize the learnable queries and temporal biases to zero and initialize the projection layers to preserve the input features.
Consequently, the attention weights are uniform at the beginning of training, $\alpha_i^{(m)}=1/r_P$, such that the attention aggregation initially reduces to temporal mean pooling before gradually learning content-adaptive temporal weighting.

\vspace{-2mm}
\paragraph{V-RAE Decoder.}
The V-RAE decoder reconstructs the input video from the compressed latent representation,
$\hat{\bm{X}}=D(\bm{Z})$.
As illustrated in Figure~\ref{fig:V-RAE-overview}, it adopts the lightweight Transformer-decoder architecture of MAE~\citep{he2022mae}, consisting of a stack of ViT blocks~\citep{dosovitskiy2021vit}.
We extend RoPE~\citep{su2021roformer} to the temporal and two spatial dimensions, providing each latent token with explicit spatiotemporal positional information.
The decoder attention pattern follows the temporal context available from the frozen representation encoder.
For DINOv3-L, SigLIP2-L, and EUPE-B, which encode frames independently, we use \textit{chunk-wise causal self-attention}: each temporal latent step and its associated spatial tokens form a chunk, tokens within a chunk interact bidirectionally, and each chunk can attend only to itself and preceding chunks.
For V-JEPA~2.1-L, the encoder representations already aggregate information with non-causal full spatiotemporal attention. We therefore use full self-attention over all spatiotemporal latent tokens in its decoder, allowing every latent chunk to attend to both preceding and subsequent chunks.
Finally, we modify the unpatchify layer to account for the overall temporal compression ratio $r_{\mathrm{all}}$.
Instead of reconstructing a single frame, each latent time step is mapped to $r_{\mathrm{all}}$ consecutive frames.
This multi-frame decoding formulation accommodates arbitrary combinations of native temporal compression in the representation encoder and additional compression introduced by the temporal pooling module.

\vspace{-2mm}
\subsection{V-RAE Learning Objective}
\vspace{-2mm}
The representation encoder remains frozen throughout V-RAE training, and only the temporal pooling module and decoder are optimized.
The training objective combines an $L_1$ reconstruction loss, a Learned Perceptual Image Patch Similarity (LPIPS) loss $\mathcal{L}_{\mathrm{LPIPS}}$~\citep{zhang2018lpips}, and an adversarial loss $\mathcal{L}_{\mathrm{GAN}}$~\citep{goodfellow2014gan}.
Following prior visual-tokenizer designs~\citep{lu2025atoken,agarwal2025cosmos}, we apply the Gram-matrix loss $\mathcal{L}_{\mathrm{Gram}}$, which encourages the reconstructed features to match the statistics of the target features.
The $L_1$ and adversarial losses are computed over all reconstructed frames.
The overall training objective is
\begin{equation}
\mathcal{L}_{\mathrm{recon}}
=
\lambda_1\mathcal{L}_1
+\lambda_{\mathrm{lpips}}\mathcal{L}_{\mathrm{LPIPS}}
+\lambda_{\mathrm{gan}}\mathcal{L}_{\mathrm{GAN}}
+\lambda_{\mathrm{gram}}\mathcal{L}_{\mathrm{Gram}}\;.
\end{equation}

\subsection{V-RAE for Generation}
\vspace{-2mm}
After reconstruction training, we freeze all V-RAE parameters and train a DiT~\citep{peebles2023dit} in the resulting latent space using rectified flow~\citep{liu2023rectifiedflow}.
To accommodate the high dimensionality of the V-RAE latent space, we follow \citet{zheng2025rae} and apply a dimension-dependent shift to the noise schedule.
Specifically, we first sample $\tau\sim\operatorname{LogitNormal}(0,1)$ and transform it as
\begin{equation}
\hat{\tau} = \frac{s \cdot \tau}{1+(s-1) \cdot \tau}\;\;,
\qquad
s = \sqrt{\frac{T_Z \cdot h \cdot w \cdot C}{n}}\;\;,
\label{eq:time-shift}
\end{equation}
where $n=4096$ is the reference latent dimension.
Unlike the corresponding shift used for image RAEs, our scaling factor explicitly accounts for the temporal latent length $T_Z$.
Given a clean latent $\bm{Z}$ and Gaussian noise $\bm{\epsilon}$, we construct the noisy latent as
\begin{equation}
    \bm{Z}_{\hat{\tau}}=(1-\hat{\tau}) \cdot \bm{Z}+\hat{\tau} \cdot \bm{\epsilon}\;.
\end{equation}
We adopt a clean-latent prediction parameterization, in which the DiT directly predicts $\widehat{\bm{Z}}_{\theta}$
from $\bm{Z}_{\hat{\tau}}$ and $\hat{\tau}$.
The clean-latent prediction is converted into a velocity estimate as
\begin{equation}
\widehat{\bm{v}}_{\theta}
=
\frac{\bm{Z}_{\hat{\tau}}-\widehat{\bm{Z}}_{\theta}
}{\max\!\left(\hat{\tau},t_\epsilon\right)}\;,
\end{equation}
where $t_\epsilon=0.05$ clamps the velocity-conversion denominator. The corresponding rectified flow target is
$\bm{v}=(\bm{Z}_{\hat{\tau}}-\bm{Z})/\max(\hat{\tau},t_\epsilon)$, which reduces to $\bm{\epsilon}-\bm{Z}$ when $\hat{\tau}\geq t_\epsilon$.
Following RAEv2~\citep{singh2026raev2}, rather than using a REPA loss~\citep{yu2024repae}, we attach an auxiliary clean-latent prediction head to the output of the $8$-th transformer block.
Let
$\mathcal{L}_{\mathrm{RF}}^{\mathrm{full}}$
and
$\mathcal{L}_{\mathrm{RF}}^{\mathrm{base}}$
denote the rectified flow losses computed from the final and auxiliary prediction heads, respectively.
The overall training objective is
\begin{equation}
\mathcal{L}_{\mathrm{DiT}}
=
\mathcal{L}_{\mathrm{RF}}^{\mathrm{full}}
+
\mathcal{L}_{\mathrm{RF}}^{\mathrm{base}}\;.
\end{equation}
During inference, the auxiliary branch is retained to provide internal guidance during sampling.
We solve the rectified flow trajectory using a 100-step Euler sampler and decode the generated latent sequence into video frames using the frozen V-RAE decoder.

\begin{table}[t]
\centering
\begin{threeparttable}
\caption{Reconstruction performance of video tokenizers on UCF101 and K600. \textbf{Bold} and \underline{underlined} values indicate the best and second-best results across all methods, respectively.}
\label{tab:reconstruction}
\vspace{1pt}
\scriptsize
{\setlength{\tabcolsep}{2pt}\renewcommand{\arraystretch}{1.5}
  \begin{tabular*}{\linewidth}{@{\extracolsep{\fill}}lccccccccccc@{}}
    \toprule
    \multicolumn{4}{c}{{\textbf{Tokenizer Configuration}}} &
    \multicolumn{4}{c}{{\textbf{UCF101 test}}} &
    \multicolumn{4}{c}{{\textbf{K600 val}}} \\
    \cmidrule(lr){5-8} \cmidrule(l){9-12}
    \textbf{Method} & \textbf{Causal} & \textbf{Comp.} & \textbf{Dim.} &
    \textbf{LPIPS} $\downarrow$ & \textbf{PSNR} $\uparrow$ & \textbf{SSIM} $\uparrow$ & \textbf{rFVD} $\downarrow$ &
    \textbf{LPIPS} $\downarrow$ & \textbf{PSNR} $\uparrow$ & \textbf{SSIM} $\uparrow$ & \textbf{rFVD} $\downarrow$ \\
    \midrule
    \rowcolor{reconbenchgroup}
    \multicolumn{12}{c}{\textbf{Large-Scale Pretrained Video Tokenizers}} \\
    Wan2.1 VAE & \cmark & $4\times8\times8$ & 16 & \secondbest{0.068} & 34.09 & 0.937 & \best{6.05} & \secondbest{0.052} & 35.23 & 0.946 & 3.58 \\
    Wan2.2 VAE & \cmark & $4\times16\times16$ & 48 & \best{0.067} & 34.16 & 0.942 & 12.20 & \best{0.049} & \secondbest{35.73} & \secondbest{0.951} & 4.76 \\
    HunyuanVideo VAE & \cmark & $4\times8\times8$ & 16 & 0.069 & \best{35.14} & \best{0.950} & 7.73 & 0.056 & \best{36.31} & \best{0.954} & 4.38 \\
    CogVideoX VAE & \cmark & $4\times8\times8$ & 16 & 0.081 & \secondbest{34.22} & \secondbest{0.943} & 14.53 & 0.069 & 34.82 & 0.946 & 8.11 \\
    Cosmos-0.1 (CV4x8x8) & \cmark & $4\times8\times8$ & 16 & 0.109 & 32.57 & 0.925 & 17.01 & 0.100 & 33.55 & 0.928 & 9.88 \\
    AToken & \xmark & $4\times16\times16$ & 48 & 0.089 & 33.11 & 0.934 & 8.17 & 0.072 & 34.05 & 0.937 & 5.36 \\
    \midrule
    \rowcolor{reconlargegroup}
    \multicolumn{12}{c}{\textbf{Benchmark-Scale Video Tokenizers}} \\
    Open-MAGVIT2 & \cmark & $4\times8\times8$ & 18 & 0.140 & 27.74 & 0.867 & 36.21 & 0.147 & 27.74 & 0.859 & 17.16 \\
    OmniTokenizer VAE & \cmark & $4\times8\times8$ & 8 & 0.132 & 29.59 & 0.908 & 28.86 & 0.139 & 27.75 & 0.899 & 15.35 \\
    LARP-L-long & \cmark & 1024 tokens & 16 & 0.233 & 25.32 & 0.799 & 142.29 & 0.236 & 26.00 & 0.801 & 125.00 \\
    \midrule
    \rowcolor{proberaegroup}
    \multicolumn{12}{c}{\textbf{V-RAE Series (Ours)}} \\
    \textbf{V-RAE} (DINOv3-L) & \cmark & $4\times16\times16$ & 1024 & 0.111 & 27.76 & 0.858 & \secondbest{6.12} & 0.104 & 27.96 & 0.850 & \secondbest{2.76} \\
    \textbf{V-RAE} (EUPE-B) & \cmark & $4\times16\times16$ & 768 & 0.122 & 26.76 & 0.829 & 8.05 & 0.114 & 27.30 & 0.829 & 3.36 \\
    \textbf{V-RAE} (SigLIP2-L) & \cmark & $4\times16\times16$ & 1024 & 0.142 & 26.41 & 0.815 & 9.83 & 0.134 & 27.05 & 0.816 & 3.38 \\
    \textbf{V-RAE} (V-JEPA2.1-L) & \xmark & $4\times16\times16$ & 1024 & 0.121 & 27.82 & 0.856 & 6.65 & 0.110 & 28.48 & 0.858 & \best{2.13} \\
    \bottomrule
  \end{tabular*}
}
\begin{tablenotes}[flushleft]
\scriptsize
\item[*]\emph{Evaluation protocol.} We sample one $256\times256$ clip from the beginning of each evaluation video with a temporal stride of 3. V-RAE, AToken, and LARP-L-long use 16-frame inputs, while causal video VAEs with an independently encoded first frame use 17-frame inputs.
\end{tablenotes}
\end{threeparttable}
\end{table}

\vspace{-2mm}
\section{Experimental Settings}
\vspace{-2mm}

\paragraph{Datasets.}
Our main reconstruction and class-conditional generation benchmarks are UCF101~\citep{soomro2012ucf101} and Kinetics-600 (K600)~\citep{carreira2018kinetics}.
For reconstruction, all V-RAE variants are trained on the union of the UCF101 and K600 training sets and evaluated on the UCF101 test set and K600 validation set, respectively.
For class-conditional generation, we train a separate DiT on each dataset using 20-frame clips sampled at an interval of 3.
Following Latte~\citep{ma2025latte}, we evaluate UCF101 generation using 2,048 generated samples; for K600, the evaluation population and covariance statistics follow the selected FVD implementation.
We additionally conduct semantic probing on UCF101, Something-Something V2 (SSv2)~\citep{goyal2017something}, and Kinetics-400 (K400)~\citep{kay2017kinetics} to assess the semantic information retained in the learned latents.

\paragraph{Implementation.}
(1) For \textit{reconstruction}, we report LPIPS~\citep{zhang2018lpips}, PSNR, SSIM~\citep{wang2004ssim}, and rFVD.
V-RAE and AToken~\citep{lu2025atoken} are evaluated on 16-frame clips at $256\times256$, whereas causal VAEs that encode the first frame separately are evaluated on 17-frame clips at the same resolution.
Unlike LARP~\citep{wang2025larp}, which reports results on the training split, we extract evaluation clips from the beginning of videos in the UCF101 test set and K600 validation set.
Clips are sampled at a temporal interval of 3 to expose larger motion changes, given FVD's known bias toward frame-level appearance over temporal realism~\citep{ge2024contentbiasfvd}, while matching the Latte preprocessing protocol~\citep{ma2025latte} used in our generation experiments.
This shared protocol supports the direct analysis in Section~\ref{sec:reconstruction-vs-generation} of the relation between reconstruction FVD (rFVD) and generation FVD (gFVD), both derived from Fr\'echet Video Distance~\citep{unterthiner2018fvd}.
For the DINOv3- and EUPE-based V-RAE variants, we initialize the decoder from the corresponding RAEv2 checkpoint~\citep{singh2026raev2}; the SigLIP2- and V-JEPA~2.1-based decoders are trained from scratch.

(2) For class-conditional \textit{generation}, all tokenizer latent spaces use the same DiT backbone and spatial resolution.
We adjust patchification according to each autoencoder's compression ratio to maintain a fixed DiT sequence length of $1{,}280$ tokens.
Causal VAEs with a separately encoded first frame generate 17 frames directly, whereas V-RAE and AToken generate 20 frames and retain the first 17 for evaluation.
Generation quality is measured by gFVD at $256\times256$.
Detailed model, training, and evaluation configurations are provided in Appendix~\ref{app:implementation}.

\vspace{-2mm}
\paragraph{Baselines.}
We compare V-RAE with large-scale pretrained video autoencoders and tokenizers, including Wan2.1 VAE and Wan2.2 VAE~\citep{wan2025wan}, HunyuanVideo VAE~\citep{kong2024hunyuanvideo}, CogVideoX VAE~\citep{yang2025cogvideox}, Cosmos VAE~\citep{agarwal2025cosmos}, AToken~\citep{lu2025atoken}, MAGVIT-v2~\citep{yu2024magvitv2,luo2024openmagvit2}, OmniTokenizer~\citep{wang2024omnitokenizer}, and LARP~\citep{wang2025larp}.
Under the matched DiT setup described above, we compare both generation quality and convergence speed.
We also include frame-wise RAEv2~\citep{singh2026raev2} to isolate the effect of temporal modeling.
Finally, we compare different temporal-pooling designs to study their effects on reconstruction quality and latent semantic quality.

\begin{figure}[!t]
    \centering
    \includegraphics[width=0.99\linewidth]{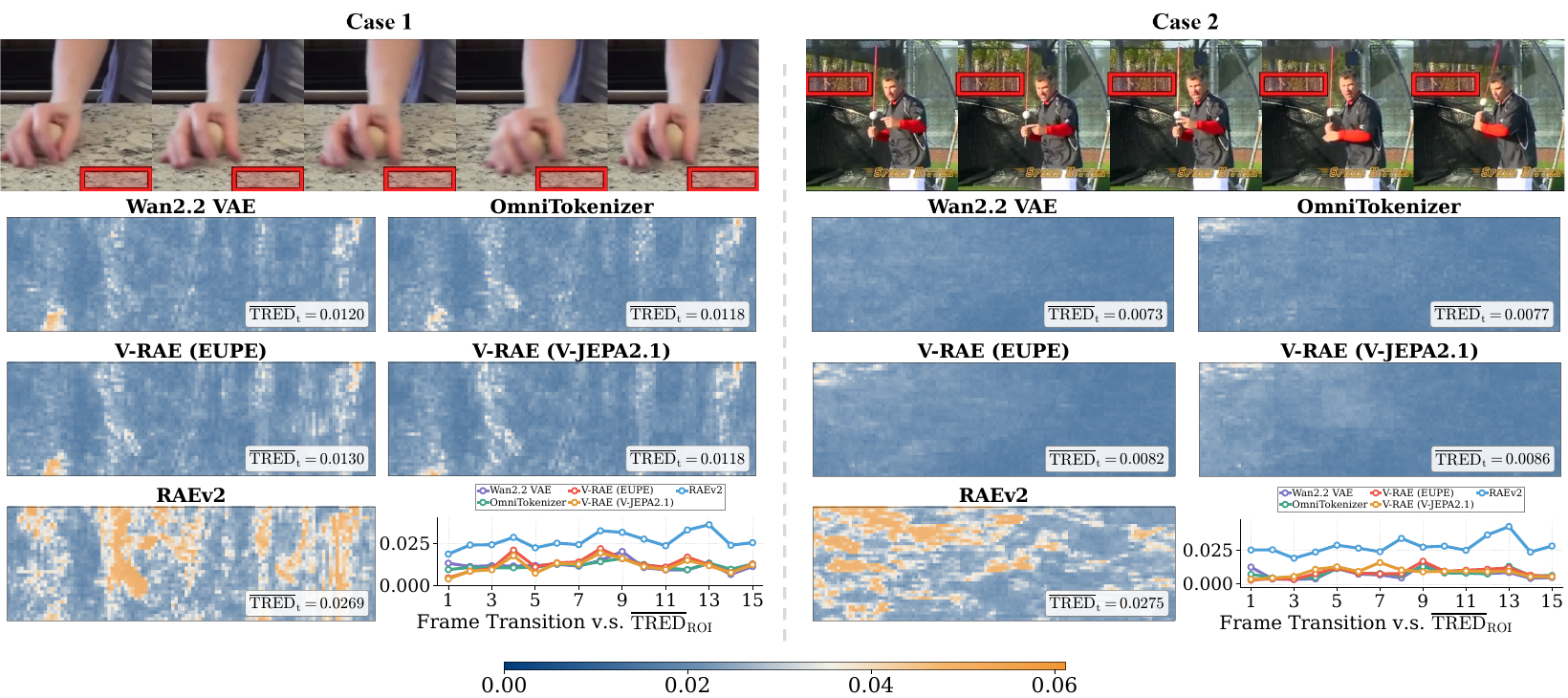}
    \vspace{-2mm}
    \caption{Temporal Reconstruction Error Difference (TRED) for RAEv2, Wan2.2 VAE, OmniTokenizer, and V-RAE on K600 validation examples. The heatmaps visualize temporally averaged TRED within high-frequency regions of interest, while the curves show spatially averaged TRED over consecutive frame transitions.}
    \label{fig:reconstruction-visualization}
    \vspace{-2mm}
\end{figure}

\vspace{-2mm}
\section{Results and Discussion}
\vspace{-2mm}
\subsection{Reconstructing Videos from Semantic Latents}

\paragraph{Overall Reconstruction Performance.}
We first compare V-RAE with both large-scale pretrained and benchmark-scale video tokenizers.
As shown in Table~\ref{tab:reconstruction}, V-RAE instantiated with DINOv3-L achieves an rFVD of 6.12 on UCF101, ranking second only to Wan2.1 VAE at 6.05.
On K600, V-RAE with V-JEPA~2.1-L achieves an rFVD of 2.13, improving over the strongest video VAE baseline at 3.58 by 40.5\%.
Moreover, all four V-RAE variants outperform the strongest evaluated large-scale video VAE baseline on K600 and consistently outperform Open-MAGVIT2, OmniTokenizer, and LARP on both datasets.
This comparison shows that representations extracted by frozen VFMs can form effective latent spaces for video reconstruction across datasets.

Comparing different representation encoders, the best reconstruction encoder is dataset-dependent: DINOv3-L performs best on UCF101, whereas V-JEPA~2.1-L performs best on K600.
The strong results of all four variants indicate that V-RAE's reconstruction capability is not specific to a single encoder or pretraining paradigm.
They further suggest that both frame-wise semantic representations and video-native spatiotemporal representations provide effective reconstruction latents, while their relative advantages depend on the target data distribution.

Because V-RAE uses a frozen semantic encoder that is not optimized specifically for pixel-level reconstruction, it underperforms some large-scale video VAEs in terms of LPIPS, PSNR, and SSIM.
Despite this, V-RAE remains highly competitive in rFVD.
This result highlights a distinction between frame-level fidelity and video-level distributional similarity: stronger pixel-wise or perceptual reconstruction does not necessarily translate into a lower distributional discrepancy in spatiotemporal feature space.
More importantly, it shows that high-quality video reconstruction can be achieved without re-optimizing the representation encoder for low-level pixel recovery.

To examine whether the improvement in rFVD is also reflected in local temporal stability, we introduce \emph{Temporal Reconstruction Error Difference} (TRED), a diagnostic metric that quantifies frame-to-frame fluctuations in reconstruction error.
Let $\mathbf{x}_t$ and $\widehat{\mathbf{x}}_t$ denote the ground-truth and reconstructed frames at time $t$, respectively, and let $\Omega_{\mathrm{ROI}}$ denote a high-frequency region of interest selected from a K600 validation video.
For each spatial location $\mathbf{p}=(h,w)\in\Omega_{\mathrm{ROI}}$, we define the channel-averaged reconstruction error as $e_t(\mathbf{p})=\frac{1}{C}\sum_{c=1}^{C}|x_{t,c}(\mathbf{p})-\hat{x}_{t,c}(\mathbf{p})|$.
We summarize its temporal variation using a time-averaged spatial heatmap and a spatially averaged temporal curve:
\begin{equation}
\begin{aligned}
\overline{\mathrm{TRED}}_{\mathrm{t}}(\mathbf{p})
&=
\frac{1}{T-1}\sum_{t=1}^{T-1}
\left|e_{t+1}(\mathbf{p})-e_t(\mathbf{p})\right|\;,
&& \mathbf{p}\in\Omega_{\mathrm{ROI}}\;,\\
\overline{\mathrm{TRED}}_{\mathrm{ROI}}(t)
&=
\frac{1}{|\Omega_{\mathrm{ROI}}|}
\sum_{\mathbf{p}\in\Omega_{\mathrm{ROI}}}
\left|e_{t+1}(\mathbf{p})-e_t(\mathbf{p})\right|\;,
&& t=1,\ldots,T-1\;.
\end{aligned}
\end{equation}
As shown in Figure~\ref{fig:reconstruction-visualization}, both V-RAE variants exhibit markedly lower $\overline{\mathrm{TRED}}_{\mathrm{t}}$ and $\overline{\mathrm{TRED}}_{\mathrm{ROI}}$ values than the frame-wise RAEv2 baseline on the presented examples, while producing values close to those of Wan2.2 VAE and OmniTokenizer.
These observations complement the global distributional comparison provided by rFVD with a localized diagnostic of reconstruction-error stability.
In particular, they suggest that the explicit temporal modeling introduced in V-RAE reduces abrupt cross-frame error fluctuations and thereby alleviates visible reconstruction flicker.

\FloatBarrier
\noindent
\begin{minipage}[t]{0.52\linewidth}
  \vspace{0pt}
  \paragraph{The Impact of Temporal Pooling.}
  We compare V-RAE with the frame-wise RAEv2 baseline to isolate the
  contribution of temporal modeling. As shown in
  Figure~\ref{fig:temporal-modeling}, each pair uses the same frozen image VFM (DINOv3-L, SigLIP2-L, or EUPE-B), thereby controlling for the underlying semantic representation.
  RAEv2 reconstructs frames independently, whereas V-RAE couples neighboring features through temporal pooling and decodes them with a causal video decoder.
  Together with the TRED visualization in Figure~\ref{fig:reconstruction-visualization}, these results highlight the importance of temporal pooling and cross-frame self-attention for improving temporal consistency and reducing flicker and jitter in reconstructed videos.
\end{minipage}\hfill
\begin{minipage}[t]{0.44\linewidth}
  \vspace{0pt}
  \centering
  \begin{tikzpicture}
    \begin{axis}[
      width=0.92\linewidth,
      height=3.75cm,
      xmin=0, xmax=80,
      ymin=0.55, ymax=3.45,
      xtick={0,20,40,60,80},
      ytick={1,2,3},
      yticklabels={EUPE-B,SigLIP2-L,DINOv3-L},
      axis x line*=bottom,
      axis y line*=left,
      y axis line style={draw=none},
      ytick style={draw=none},
      tick align=outside,
      xmajorgrids,
      grid style={draw=black!12, line width=0.3pt},
      tick label style={font=\tiny},
      yticklabel style={font=\scriptsize, text=black},
      xlabel={UCF101 rFVD $\downarrow$},
      xlabel style={font=\scriptsize, yshift=2pt},
      legend style={
        draw=none,
        fill=none,
        font=\scriptsize,
        at={(0.5,1.02)},
        anchor=south,
        legend columns=2,
        /tikz/every even column/.append style={column sep=5pt}
      },
      clip=false
    ]
      \addplot[draw=chartbaseline!60, line width=1.1pt, forget plot]
        coordinates {(27.28,1) (8.05,1)};
      \addplot[draw=chartbaseline!60, line width=1.1pt, forget plot]
        coordinates {(74.65,2) (9.83,2)};
      \addplot[draw=chartbaseline!60, line width=1.1pt, forget plot]
        coordinates {(16.27,3) (6.12,3)};

      \addplot[
        only marks, mark=*, mark size=2.5pt,
        draw=chartbaseline, fill=chartbaseline,
        point meta=x,
        nodes near coords={\pgfmathprintnumber[fixed,precision=2]{\pgfplotspointmeta}},
        every node near coord/.append style={font=\tiny, anchor=south west, text=chartbaseline}
      ] coordinates {(27.28,1) (74.65,2) (16.27,3)};
      \addlegendentry{RAEv2}

      \addplot[
        only marks, mark=*, mark size=2.5pt,
        draw=chartours, fill=chartours,
        point meta=x,
        nodes near coords={\pgfmathprintnumber[fixed,precision=2]{\pgfplotspointmeta}},
        every node near coord/.append style={font=\tiny\bfseries, anchor=north east, text=chartours}
      ] coordinates {(8.05,1) (9.83,2) (6.12,3)};
      \addlegendentry{V-RAE}
    \end{axis}
  \end{tikzpicture}
  \vspace{-5pt}
  
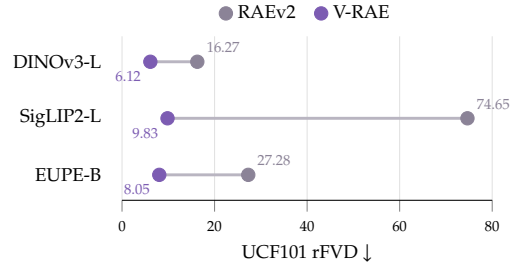
\captionof{figure}{Effect of temporal modeling on UCF101 reconstruction; lower rFVD is better.}
  \label{fig:temporal-modeling}
\end{minipage}

\begin{figure}[!t]
  \centering
  \begin{minipage}[b]{0.73\linewidth}
    \centering
    \begin{tikzpicture}
      \begin{axis}[
        width=\linewidth,
        height=5.35cm,
        xmin=5.80, xmax=7.10,
        ymin=76.5, ymax=93.4,
        xtick={5.9,6.2,6.5,6.8,7.1},
        ytick={78,82,86,90},
        xlabel={rFVD $\downarrow$},
        ylabel={Probe Accuracy (\%) $\uparrow$},
        axis lines=left,
        tick align=outside,
        tick label style={font=\scriptsize, text=black!70},
        label style={font=\small},
        axis line style={draw=black!45},
        tick style={draw=black!35},
        grid=major,
        grid style={draw=black!8, line width=0.4pt},
        clip=false
      ]
        \path[fill=chartours!6, draw=none]
          (axis cs:5.80,87.0) rectangle (axis cs:7.10,93.4);
        \node[anchor=north west, font=\tiny\bfseries, text=chartours!75!black]
          at (axis cs:5.84,93.05) {better semantic preservation};

        \addplot[
          chartours!70!black,
          densely dashed,
          line width=0.8pt,
          forget plot
        ] coordinates {(5.91,78.26) (6.12,89.13)};
        \node[font=\tiny, text=chartours!70!black, fill=white, inner sep=1.2pt]
          at (axis cs:6.22,83.3) {Pareto frontier};

        \addplot[
          only marks, mark=*, mark size=2.8pt,
          draw=chartbaseline, fill=chartbaseline,
          forget plot
        ] coordinates {(5.91,78.26)};
        \node[anchor=south west, align=left, font=\scriptsize, text=chartbaseline!75!black]
          at (axis cs:5.94,78.65)
          {\textbf{Convolutional}\\[-1pt]\tiny 5.91 / 78.26 / 4M};

        \addplot[
          only marks, mark=*, mark size=2.8pt,
          draw=chartbaseline!75!black, fill=chartbaseline!75,
          forget plot
        ] coordinates {(6.87,84.37)};
        \node[anchor=east, align=right, font=\scriptsize, text=chartbaseline!75!black]
          at (axis cs:6.845,84.37)
          {\textbf{Q-Former} \textit{(dominated)}\\[-1pt]\tiny 6.87 / 84.37 / 67M};

        \addplot[
          only marks, mark=*, mark size=5.0pt,
          draw=chartours!25, fill=chartours!12,
          forget plot
        ] coordinates {(6.12,89.13)};
        \addplot[
          only marks, mark=*, mark size=2.9pt,
          draw=chartours!85!black, fill=chartours,
          forget plot
        ] coordinates {(6.12,89.13)};
        \node[
          anchor=south west, align=left, font=\scriptsize,
          text=chartours!75!black, fill=chartours!8,
          rounded corners=1.5pt, inner sep=2.2pt
        ] at (axis cs:6.15,89.35)
          {\textbf{Temporal Attention (default)}\\[-1pt]\tiny 6.12 / 89.13 / 3M};
      \end{axis}
      \draw[black!45, line width=0.7pt]
        ([xshift=-5pt,yshift=-3pt]current axis.south east) -- ++(3pt,7pt);
      \draw[black!45, line width=0.7pt]
        ([xshift=0pt,yshift=-3pt]current axis.south east) -- ++(3pt,7pt);
    \end{tikzpicture}
  \end{minipage}\hfill
  \begin{minipage}[b]{0.23\linewidth}
    \centering
    \begin{tikzpicture}
      \begin{axis}[
        width=\linewidth,
        height=5.35cm,
        xmin=90.5, xmax=97.0,
        ymin=76.5, ymax=93.4,
        xtick={94.25},
        ytick={78,82,86,90},
        yticklabels={,,,},
        xlabel={rFVD $\downarrow$},
        axis x line*=bottom,
        axis y line=none,
        tick align=outside,
        tick label style={font=\scriptsize, text=black!70},
        label style={font=\small},
        axis line style={draw=black!45},
        tick style={draw=black!35},
        ymajorgrids=true,
        grid style={draw=black!8, line width=0.4pt},
        clip=false
      ]
        \path[fill=chartours!6, draw=none]
          (axis cs:90.5,87.0) rectangle (axis cs:97.0,93.4);
        \addplot[
          only marks, mark=*, mark size=2.8pt,
          draw=chartbaseline, fill=chartbaseline,
          forget plot
        ] coordinates {(94.25,91.82)};
        \node[anchor=north, align=center, font=\scriptsize, text=chartbaseline!75!black]
          at (axis cs:94.25,91.45)
          {\textbf{Mean}\\[-1pt]\tiny 94.25 / 91.82 / 0};
      \end{axis}
      \draw[black!45, line width=0.7pt]
        ([xshift=-3pt,yshift=-3pt]current axis.south west) -- ++(3pt,7pt);
      \draw[black!45, line width=0.7pt]
        ([xshift=2pt,yshift=-3pt]current axis.south west) -- ++(3pt,7pt);
    \end{tikzpicture}
  \end{minipage}
  \vspace{-3pt}
  \caption{Reconstruction-semantics trade-off of temporal pooling designs on the UCF101 test set. Point labels report rFVD / probing accuracy / parameters; the horizontal axis is broken to retain resolution among the learnable designs.}
  \label{fig:temporal-pooling-ablation}
  \vspace{-5pt}
\end{figure}
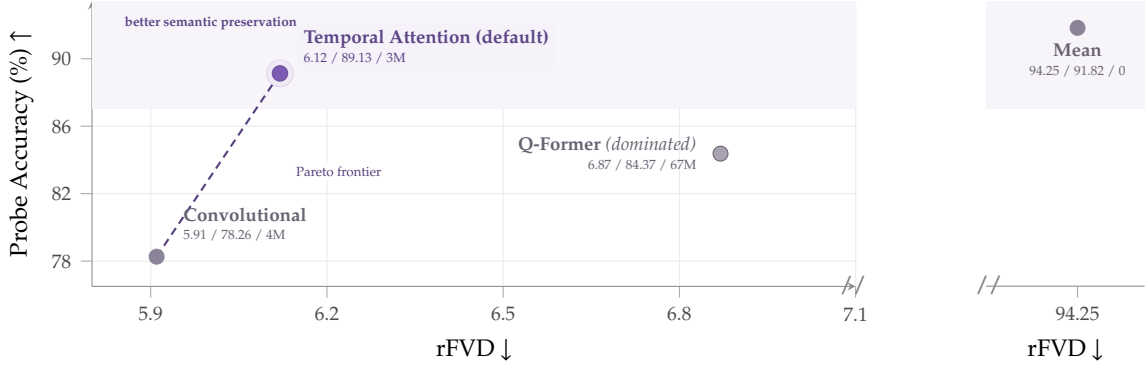

\paragraph{Comparison of Temporal Pooling Designs.}
To isolate the effect of temporal pooling, we keep the frozen representation encoder, video decoder, and training recipe fixed, and vary only the pooling mechanism.
Detailed structures of the four pooling designs are provided in
Appendix~\ref{app:pooling-designs}.
The trade-offs in Figure~\ref{fig:temporal-pooling-ablation} can therefore be attributed primarily to the temporal aggregation strategy.
Mean pooling introduces no additional parameters and achieves the highest probing accuracy (91.82\%), suggesting that uniform averaging largely preserves the semantic structure inherited from the pretrained encoder.
However, it yields substantially worse reconstruction fidelity, with a 94.25 rFVD, as it cannot adapt its aggregation to content-dependent temporal variations.
Convolutional pooling substantially reduces rFVD to 5.91 through learnable local temporal mixing, but its probing accuracy drops to 78.26\%, indicating that improved reconstruction comes at the cost of semantic preservation.
Increasing the pooling capacity does not resolve this trade-off: Q-former pooling uses 67M parameters, but neither improves reconstruction (6.87 rFVD) nor fully preserves semantic information (84.37\% probing accuracy).
Additional capacity alone is therefore insufficient to balance reconstruction and semantic preservation.
In contrast, temporal attention pooling adaptively weights features according to their temporal content.
With only 3M parameters, it achieves 6.12 rFVD while retaining 89.13\% probing accuracy, providing a substantially better reconstruction-semantics balance than the other pooling strategies.
We therefore adopt temporal attention pooling as the default configuration in all remaining experiments.

\begin{table}[!t]
  \caption{Top-1 probing accuracy of RAE- and VAE-based video tokenizers. Parentheses indicate the change from the corresponding frozen model.}
  \label{tab:semantic-probing}
  \vspace{1pt}
  \centering
  \footnotesize
  {\setlength{\tabcolsep}{2.5pt}\renewcommand{\arraystretch}{1.16}
  \begin{tabularx}{\linewidth}{@{}
    >{\raggedright\arraybackslash}X
    >{\centering\arraybackslash}p{0.105\linewidth}
    *{6}{>{\centering\arraybackslash}p{0.0925\linewidth}}
  @{}}
    \toprule
    {\textbf{Model}} &
    \multirow{3}{=}{\centering\scriptsize\hyphenpenalty=10000\exhyphenpenalty=10000\bfseries Temporal Compression Ratio\par} &
    \multicolumn{2}{c}{{\textbf{UCF101} $\uparrow$}} &
    \multicolumn{2}{c}{{\textbf{SSv2} $\uparrow$}} &
    \multicolumn{2}{c}{{\textbf{K400} $\uparrow$}} \\
    \cmidrule(lr){3-4} \cmidrule(lr){5-6} \cmidrule(l){7-8}
    \textit{Probe Type} & & \multicolumn{2}{c}{\textit{Linear Probe}} & \multicolumn{2}{c}{\textit{Attentive Probe}} & \multicolumn{2}{c}{\textit{Attentive Probe}} \\
    & & \textbf{Single} & \textbf{TTA} & \textbf{Single} & \textbf{TTA} & \textbf{Single} & \textbf{TTA} \\
    \midrule
    \rowcolor{proberaegroup}
    \multicolumn{8}{c}{\textbf{RAE-based Video Autoencoders}} \\
    DINOv3-L & $1\times$ & 91.84 & 91.84 & 69.92 & 70.21 & 86.32 & 87.15 \\
    \rowcolor{probevraerow}
    \textbf{V-RAE} (DINOv3-L) & $4\times$ & 89.13\reldrop{-2.71} & 88.94\reldrop{-2.90} & 66.55\reldrop{-3.37} & 67.19\reldrop{-3.02} & 83.12\reldrop{-3.20} & 84.36\reldrop{-2.79} \\
    \addlinespace[2pt]
    EUPE-B & $1\times$ & 93.86 & 93.78 & 67.17 & 67.64 & 83.53 & 84.21 \\
    \rowcolor{probevraerow}
    \textbf{V-RAE} (EUPE-B) & $4\times$ & 90.16\reldrop{-3.70} & 89.93\reldrop{-3.85} & 65.67\reldrop{-1.50} & 66.21\reldrop{-1.43} & 82.21\reldrop{-1.32} & 83.31\reldrop{-0.90} \\
    \addlinespace[2pt]
    SigLIP2-L & $1\times$ & 94.05 & 93.55 & 68.54 & 68.80 & 86.00 & 86.93 \\
    \rowcolor{probevraerow}
    \textbf{V-RAE} (SigLIP2-L) & $4\times$ & 90.92\reldrop{-3.13} & 90.88\reldrop{-2.67} & 65.39\reldrop{-3.15} & 66.08\reldrop{-2.72} & 82.56\reldrop{-3.44} & 83.91\reldrop{-3.02} \\
    \addlinespace[2pt]
    V-JEPA 2.1-L & $2\times$ & 93.02 & 92.83 & 76.58 & 77.11 & 84.29 & 85.20 \\
    \rowcolor{probevraerow}
    \textbf{V-RAE} (V-JEPA 2.1-L) & $4\times$ & 86.65\reldrop{-6.37} & 86.65\reldrop{-6.18} & 72.91\reldrop{-3.67} & 73.50\reldrop{-3.61} & 80.70\reldrop{-3.60} & 81.87\reldrop{-3.34} \\
    \midrule
    \rowcolor{probevaegroup}
    \multicolumn{8}{c}{\textbf{VAE-based Video Autoencoders}} \\
    Wan2.2 VAE & $4\times$ & 16.94 & 16.29 & 41.86 & 42.62 & 46.13 & 48.11 \\
    AToken & $4\times$ & 30.83 & 30.29 & 45.05 & 45.95 & 53.27 & 55.32 \\
    CogVideoX VAE & $4\times$ & 14.51 & 14.14 & 37.03 & 38.34 & 41.59 & 43.53 \\
    \bottomrule
  \end{tabularx}}
  \vspace{-6pt}
\end{table}

\paragraph{Semantic Preservation.}

We probe the temporally compressed generation tokens on UCF101, SSv2, and K400, following the DINOv3 protocol~\citep{simeoni2025dinov3}.
Representative VAE-based video tokenizers are included for comparison.

As shown in Table~\ref{tab:semantic-probing}, V-RAE latents retain substantially richer semantic information than conventional VAE latents.
V-RAE achieves 90.92\% on UCF101 with SigLIP2-L, 72.91\% on SSv2 with V-JEPA~2.1, and 83.12\% on K400 with DINOv3-L, compared with 30.83\%, 45.05\%, and 53.27\% for the strongest evaluated VAE baseline, respectively.
The gains across diverse benchmarks indicate that the compressed latents retain both appearance and motion semantics.

\begin{table}[!t]
  \centering
  \begin{threeparttable}
  \caption{Class-conditional video generation performance comparison on UCF101 and K600 using different video tokenizers and video VAEs as latent spaces.}
  \label{tab:controlled-generation}
  \vspace{1pt}
  \footnotesize
  {\setlength{\tabcolsep}{3pt}\renewcommand{\arraystretch}{1.16}
  \begin{tabularx}{\linewidth}{@{}
    >{\raggedright\arraybackslash}X
    >{\centering\arraybackslash}p{0.18\linewidth}
    >{\centering\arraybackslash}p{0.14\linewidth}
    >{\centering\arraybackslash}p{0.09\linewidth}
    *{2}{>{\centering\arraybackslash}p{0.13\linewidth}}
  @{}}
    \toprule
    \textbf{Method} &
    \textbf{Resolution, Frames} &
    \textbf{Generator} &
    \textbf{\#Tokens} &
    \textbf{UCF101 gFVD} $\downarrow$ &
    \textbf{K600 gFVD} $\downarrow$ \\
    \midrule
    \rowcolor{probevaegroup}
    \multicolumn{6}{c}{\textbf{VAE-Based Video Autoencoders}} \\
    Wan2.1 VAE & $256\times256$, 17 & DiT XL/2 & 1280 & 148.20 & 53.75 \\
    Wan2.2 VAE & $256\times256$, 17 & DiT XL/1 & 1280 & 154.64 & 52.25 \\
    HunyuanVideo VAE & $256\times256$, 17 & DiT XL/2 & 1280 & 211.53 & 53.28 \\
    CogVideoX VAE & $256\times256$, 17 & DiT XL/2 & 1280 & 159.20 & 51.83 \\
    Cosmos VAE (CV4x8x8) & $256\times256$, 17 & DiT XL/2 & 1280 & 152.70 & 41.66 \\
    AToken & $256\times256$, $20\!\to\!17$ & DiT XL/1 & 1280 & 143.00 & 46.74 \\
    \midrule
    \rowcolor{proberaegroup}
    \multicolumn{6}{c}{\textbf{V-RAE Series (Ours)}} \\
    \textbf{V-RAE} (DINOv3-L) & $256\times256$, $20\!\to\!17$ & DiT$^{\mathrm{DH}}$ XL/1 & 1280 & 131.40 & 30.09 \\
    \textbf{V-RAE} (SigLIP2-L) & $256\times256$, $20\!\to\!17$ & DiT$^{\mathrm{DH}}$ XL/1 & 1280 & 142.60 & 34.48 \\
    \textbf{V-RAE} (EUPE-B) & $256\times256$, $20\!\to\!17$ & DiT$^{\mathrm{DH}}$ XL/1 & 1280 & 125.98 & 24.77 \\
    \textbf{V-RAE} (V-JEPA 2.1-L) & $256\times256$, $20\!\to\!17$ & DiT$^{\mathrm{DH}}$ XL/1 & 1280 & \best{117.86} & \best{19.16} \\
    \bottomrule
  \end{tabularx}
  }
  \begin{tablenotes}[flushleft]
    \scriptsize
    \item[*] \emph{Note:} DiT$^{\mathrm{DH}}$ denotes dual-head prediction. Following RAE~\citep{zheng2025rae}, which reports \textbf{no benefit and sometimes degradation} for low-dimensional VAE latents, VAE baselines retain single-head prediction. All methods use 1280 latent tokens; $20\!\to\!17$ denotes generating 20 frames and cropping to 17 for evaluation.
  \end{tablenotes}
  \end{threeparttable}
  \vspace{-2mm}
\end{table}

Temporal compression also largely preserves the semantics of the original encoder features: the three image-encoder variants remain within 3.85 percentage points of their frozen encoders across all datasets and evaluation settings.
V-JEPA~2.1 shows a larger 6.37-point decrease on UCF101, but remains within 3.67 points on SSv2 and K400 while achieving the strongest V-RAE result on SSv2.
Temporal attention pooling therefore substantially reduces temporal redundancy with only modest semantic degradation.

\finding{1}{Temporal compression need not erase semantics.}{
The key is not pooling capacity but content-adaptive aggregation: an
effective temporal compressor selectively removes temporal redundancy while
preserving the semantic structure inherited from the pretrained encoder.}

\subsection{Generation from Semantic Latents}

\vspace{-1mm}
\paragraph{Overall Generation Performance.}

We then evaluate the same semantic latents as a space for class-conditional video generation.
Following the shared training configuration in Appendix~\ref{app:latent-dit}, we train a class-conditional DiT from scratch on UCF101 and K600 for each tokenizer and decode its generated latents with the corresponding decoder.
We match the latent budget at 1,280 tokens across methods, thereby isolating the effect of the latent representation under comparable generation settings.

\begin{figure}[!t]
    \vspace*{5mm}
    \centering
    \includegraphics[width=\linewidth]{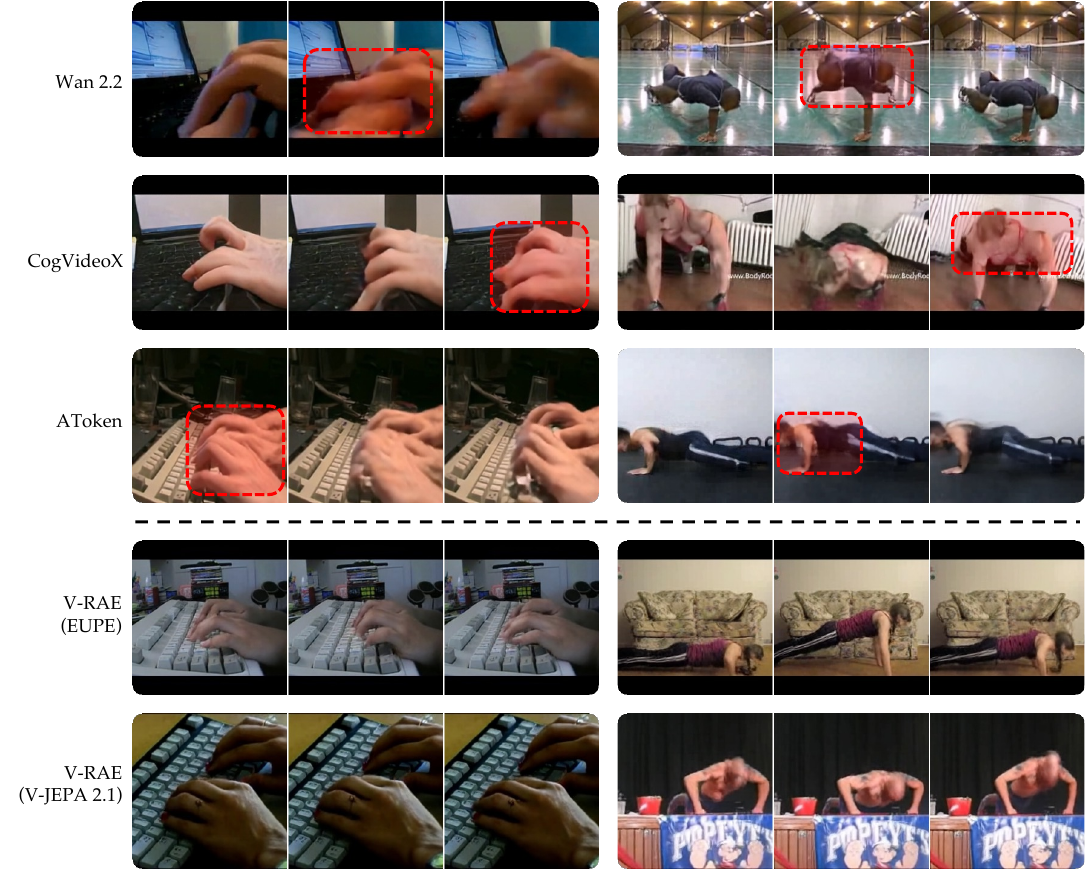}
    \caption{Qualitative class-conditional generation on UCF101.
    Dashed lines separate the baselines and V-RAE models; each strip shows three frames from one sample.
    }
    \label{fig:generation-visualization-ucf101}
    \vspace{-2mm}
\end{figure}

\begin{figure}[!t]
    \centering
    \includegraphics[width=\linewidth]{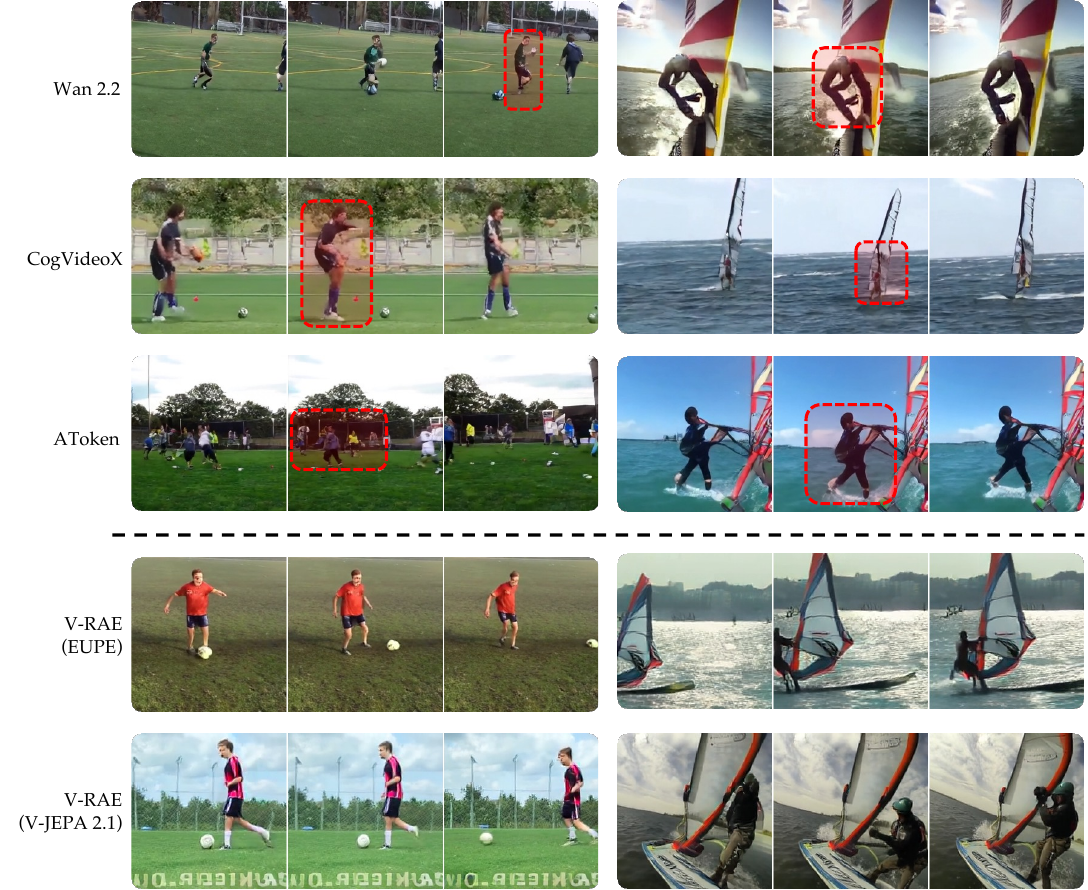}
    \caption{Qualitative class-conditional generation on Kinetics-600.
    Dashed lines separate the baselines and V-RAE models; each strip shows three frames from one sample.
    }
    \label{fig:generation-visualization-k600}
    \vspace{-2mm}
\end{figure}

Table~\ref{tab:controlled-generation} shows that all four V-RAE variants outperform conventional video tokenizers on both datasets.
V-RAE with V-JEPA~2.1 performs best, achieving 117.86 gFVD on UCF101 and 19.16 on K600, improvements of 25.14 and 22.50 points over the strongest non-V-RAE baselines.
EUPE-B also achieves 125.98 and 24.77, confirming that the advantage extends across representation encoders.
Figures~\ref{fig:generation-visualization-ucf101} and~\ref{fig:generation-visualization-k600} show improved visual integrity and cross-frame consistency; additional DINOv3-L and SigLIP2-L results appear in Appendix~\ref{app:additional-generation-results}.

\noindent
\begin{minipage}[t]{0.56\textwidth}
\vspace{0pt}
\textbf{Optimization Efficiency.}
Final gFVD captures generation quality only after training. To examine optimization efficiency, we additionally track gFVD throughout training under matched optimization settings.
Figure~\ref{fig:generation-convergence} compares V-RAE with Wan2.2 VAE on UCF101.
On UCF101, V-JEPA~2.1 and EUPE-B maintain lower gFVD than Wan2.2 VAE at every evaluated checkpoint. In particular, V-JEPA~2.1 reaches, in roughly 30K updates, a gFVD comparable to that attained by Wan2.2 VAE after 150K updates, corresponding to approximately $5\times$ faster convergence.
As shown in Figure~\ref{fig:hero}, the advantage is more pronounced on K600.
EUPE-B matches the gFVD reached by Wan2.2 VAE at 180K updates after only 30K updates.
This corresponds to $6\times$ faster convergence, while at 180K updates it outperforms Wan2.2 VAE by 24.1 gFVD points.
\end{minipage}\hfill
\begin{minipage}[t]{0.41\textwidth}
\vspace{0pt}
\centering
\includegraphics[width=\linewidth,trim=25 10 35 18,clip]{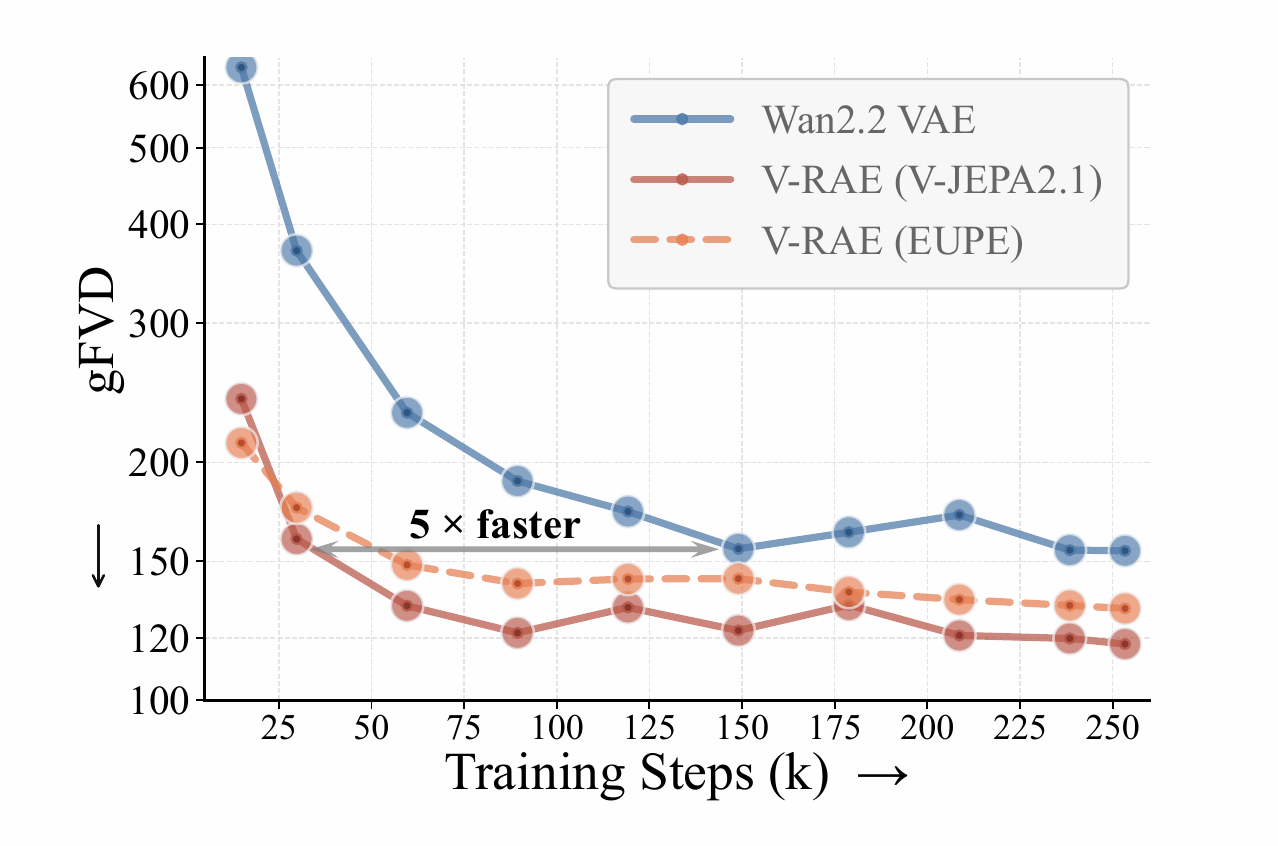}
\captionsetup{font=small,skip=3pt}
\captionof{figure}{gFVD convergence on UCF101 under matched training settings. V-RAE converges up to \(5\times\) faster than Wan2.2 VAE.}
\label{fig:generation-convergence}
\end{minipage}
\par\vspace{2pt}
These convergence results complement the final scores in Table~\ref{tab:controlled-generation}
and show that semantic latents are not only more effective generation targets but also substantially easier to optimize.
\par\vspace{8pt}

\Needspace{7\baselineskip}
\finding{2}{Semantic organization makes video generation easier to learn.}{
Semantic latents expose objects, actions, and scene structure to the generator, allowing it to focus on modeling how visual states evolve rather than rediscovering semantics and dynamics from reconstruction-oriented codes.}

\begin{figure}[!t]
    \centering
    \begin{minipage}[b]{0.53\linewidth}
        \centering
        \includegraphics[width=\linewidth]{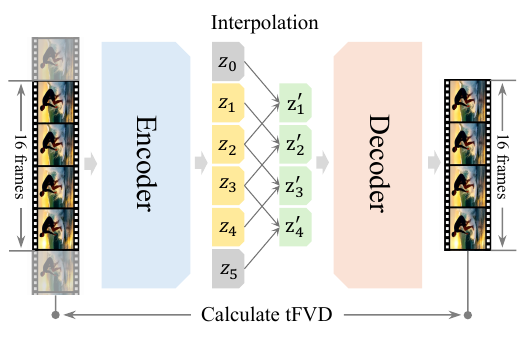}
        \vspace{-2pt}
        {\small (a) tFVD computation}
    \end{minipage}\hfill
    \begin{minipage}[b]{0.45\linewidth}
        \centering
        \includegraphics[width=\linewidth]{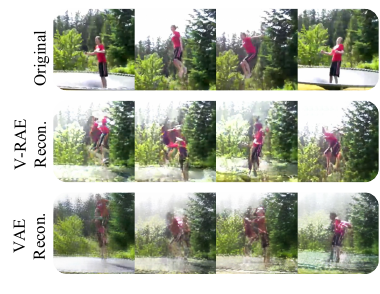}
        \vspace{-2pt}
        {\small (b) Visualization of interpolated reconstructions}
    \end{minipage}
    \caption{Overview of tFVD. (a) Each interior latent is replaced by the midpoint of its two temporal neighbors, and the interpolated sequence is decoded to compute FVD against real clips. (b) Qualitative comparison of interpolated reconstructions from V-RAE and a conventional VAE.}
    \label{fig:tfvd-interpolation}
    \vspace{-3mm}
\end{figure}

\subsection{Good Reconstruction Does Not Guarantee Good Generation}
\label{sec:reconstruction-vs-generation}

Reconstruction quality alone does not characterize whether a latent space is well suited to generative modeling.
Reconstruction evaluates the deterministic path $\bm{X}\xrightarrow{\mathcal{E}}\bm{F}\xrightarrow{\mathcal{P}}\bm{Z}\xrightarrow{\mathcal{D}}\widehat{\bm{X}}$, where the decoder receives latents encoded from real videos.
A generator faces a different challenge: it must learn the latent distribution and generate new latent samples that can be reliably decoded.
Consequently, an autoencoder may reconstruct real samples faithfully while still inducing a latent space that is difficult for a downstream generator to model.

Our results expose this discrepancy directly. On UCF101, HunyuanVideo VAE
achieves a lower rFVD than V-RAE with EUPE-B (7.73 vs.\ 8.05), but a
substantially worse gFVD (211.53 vs.\ 125.98). On K600, Wan2.1 VAE and V-RAE
with SigLIP2-L achieve similar rFVD scores (3.58 vs.\ 3.38), yet their gFVD
scores differ substantially (53.75 vs.\ 34.48). Thus, fidelity at encoded data points does not by itself reveal
how easily the surrounding latent space can be learned and decoded.

\paragraph{Temporal Fr\'echet Video Distance (tFVD).}
To assess this property without training a complete generator, we introduce
\textbf{tFVD}, a generation-oriented diagnostic that probes the local temporal
geometry of a video latent space. The complete evaluation protocol is provided
in Appendix~\ref{app:evaluation}. Given a temporally ordered latent sequence
\(\bm{Z}=[\bm{z}_0,\bm{z}_1,\ldots,\bm{z}_{L+1}]\) with \(L\) interior codes, where \(\bm{z}_0\) is the first-frame latent for causal VAEs and the first-four-frame latent for V-RAE, we replace each interior code by the midpoint
of its two temporal neighbors:
\begin{equation}
\bm{z}_t'=\frac{1}{2}\left(\bm{z}_{t-1}+\bm{z}_{t+1}\right)\;,
\qquad t=1,\ldots,L\;.
\end{equation}
We then form the perturbed latent sequence $\bm{Z}'=[\bm{z}'_1,\ldots,\bm{z}'_L]$, decode it, and compute FVD~\citep{unterthiner2018fvd} between the reconstructed clips and their corresponding real clips:
\begin{equation}
\operatorname{tFVD}
=
\operatorname{FVD}
\left(
\{\bm{X}\},
\{\mathcal{D}(\bm{Z}')\}
\right)\;.
\end{equation}
A lower tFVD indicates that local interpolations between temporally adjacent latent codes can be decoded into temporally coherent videos.

Unlike rFVD, this construction deliberately breaks the direct encode-decode path by requiring the decoder to process latent codes that are not directly produced by the encoder at corresponding time steps.
It therefore serves as a controlled stress test rather than a full generative process.
If the temporal latent trajectory is highly curved or discontinuous, its local midpoints may leave the region that decodes to natural videos, causing ghosting, flicker, or abrupt motion.
A smoother temporal latent space should instead decode these interpolated codes into coherent intermediate motion, potentially making it more tolerant to latent prediction errors during generation.

Figure~\ref{fig:tfvd-interpolation}(b) illustrates the resulting difference.
For the jumping sequence, V-RAE better preserves the subject, action trajectory, and scene structure across the interpolated frames, whereas the VAE exhibits stronger ghosting, texture corruption, and structural instability.
Although both models can reconstruct encoded inputs, their behavior between encoded temporal states is markedly different.

\begin{figure}[!t]
    \centering
    \includegraphics[width=0.96\linewidth]{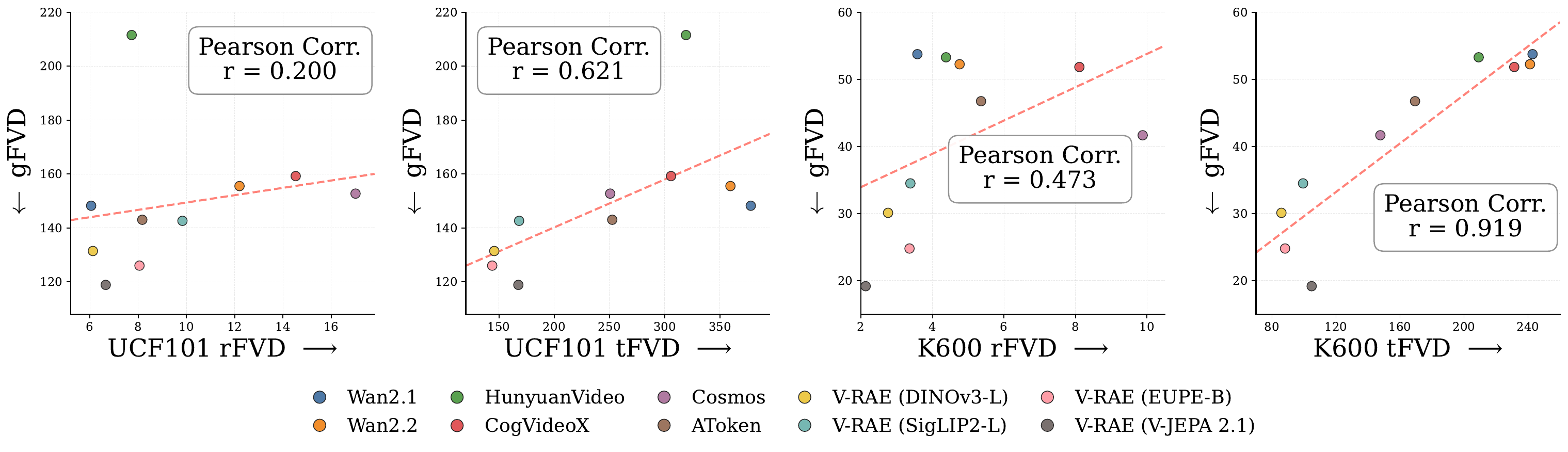}
    \vspace{-2mm}
    \caption{Metric-correlation analysis on UCF101 and K600, comparing reconstruction fidelity and temporal interpolation consistency against downstream generation quality.}
    \label{fig:metric-correlation}
    \vspace{-4mm}
\end{figure}

\vspace{-4mm}
\paragraph{Predicting Generation Performance.}
To assess whether autoencoder-level metrics predict downstream generation quality, we compare rFVD and tFVD with gFVD under matched generator settings.
As shown in Figure~\ref{fig:metric-correlation}, rFVD is substantially less correlated with gFVD: the Pearson correlations are \(r=0.200\) on UCF101 and \(r=0.473\) on K600.
In contrast, tFVD increases the correlations to \(r=0.621\) and \(r=0.919\), respectively.
The substantially stronger association across both datasets shows that temporal interpolation consistency provides a more informative estimate of generative utility than reconstruction fidelity alone.

The analysis shows that reconstruction fidelity alone is insufficient for evaluating a video tokenizer.
A generation-friendly latent space must also be semantically organized, temporally smooth, and robust to small deviations from encoded trajectories.
Together with the semantic probing and generation results, tFVD indicates that V-RAE benefits not merely from reconstructing videos with frozen representation features, but from organizing those features into a coherent latent space that is easier for a video generator to learn.

\finding{3}{Generation-friendly latents require more than reconstruction fidelity.}{
A good video latent should remain temporally smooth and decodable under prediction errors. tFVD probes this latent temporal smoothness and decoder robustness more directly than rFVD.}

\subsection{Future Video Prediction in Semantic Latent Space}
\label{sec:world-modeling}

Beyond class-conditional video generation, we investigate whether V-RAE can provide a directly decodable latent state space for visual world modeling, i.e., predicting future clips based on context.
Technically, let $\mathbf{X}^{c}$ and $\mathbf{X}^{f}$ denote the context and future clips, respectively.
The prediction pipeline is formulated as
\begin{equation}
\mathbf{Z}^{c}
=
\mathcal{P}\!\left(\mathcal{E}\!\left(\mathbf{X}^{c}\right)\right),
\qquad
\mathbf{Z}^{f}
=
\mathcal{P}\!\left(\mathcal{E}\!\left(\mathbf{X}^{f}\right)\right),
\qquad
\widehat{\mathbf{Z}}^{f}
\sim
p_{\theta}\!\left(
\mathbf{Z}^{f}\mid\mathbf{Z}^{c}
\right),
\qquad
\widehat{\mathbf{X}}^{f}
=
\mathcal{D}\!\left(\widehat{\mathbf{Z}}^{f}\right),
\label{eq:world-modeling}
\end{equation}
where $\mathbf{Z}^{c}$ and $\mathbf{Z}^{f}$ are the context and target future latents produced by the frozen representation encoder $\mathcal{E}$ followed by the temporal pooler $\mathcal{P}$, and $\widehat{\mathbf{Z}}^{f}$ is the prediction sampled from the conditional latent DiT $p_{\theta}(\cdot|\cdot)$ given $\mathbf{Z}^{c}$. The V-RAE decoder $\mathcal{D}$ then renders the predicted future clip $\widehat{\mathbf{X}}^{f}$.
This formulation jointly evaluates whether the latent space supports predictable future dynamics and whether predicted states can be decoded while preserving scene geometry, object identity, and motion trajectories.

\begin{figure}[p]
    \centering
    \includegraphics[width=0.88\linewidth]{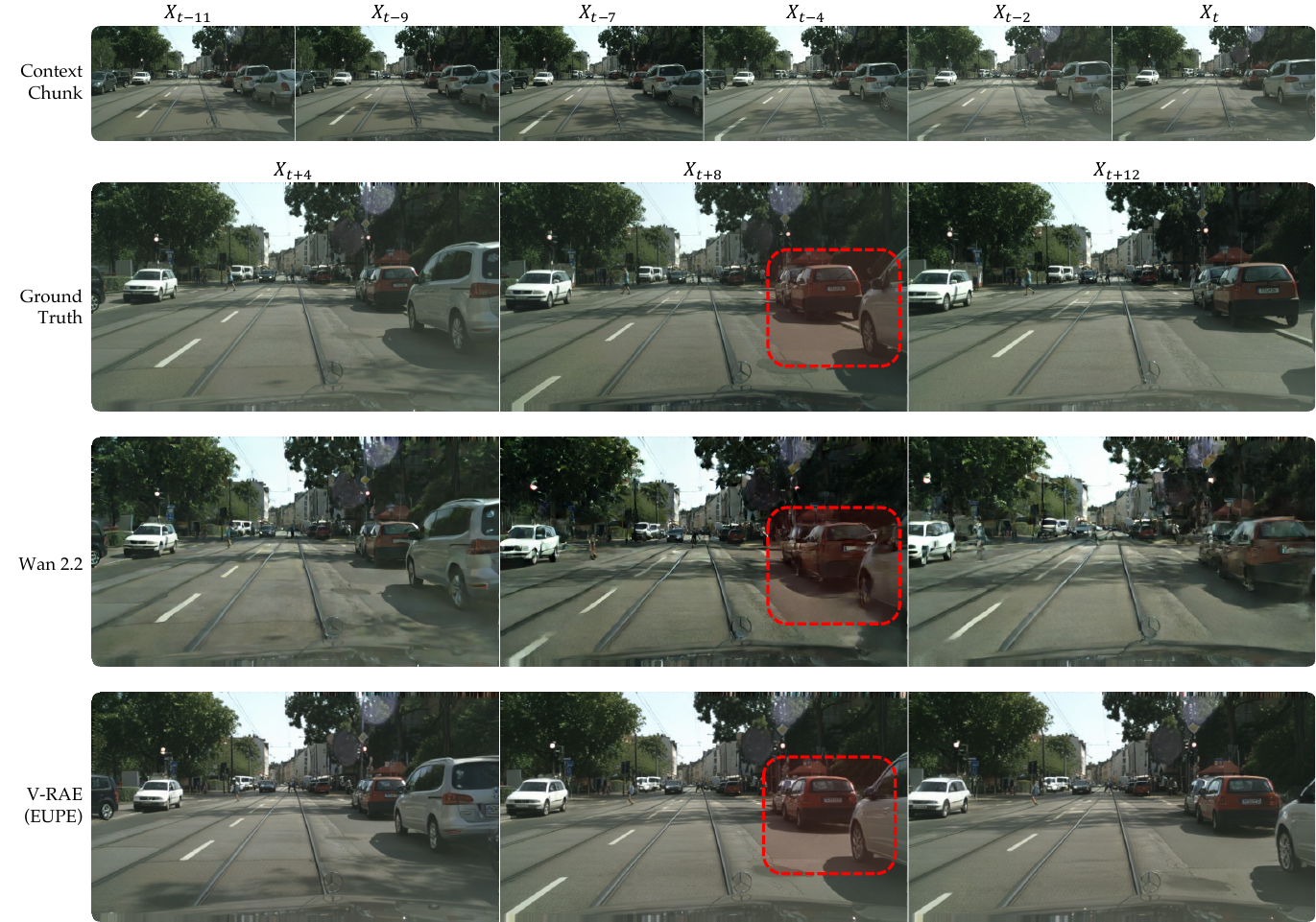}
    \par
    {\small (a) Vehicle-dominated traffic scene}\par

    \vspace{1mm}
    \includegraphics[width=0.88\linewidth]{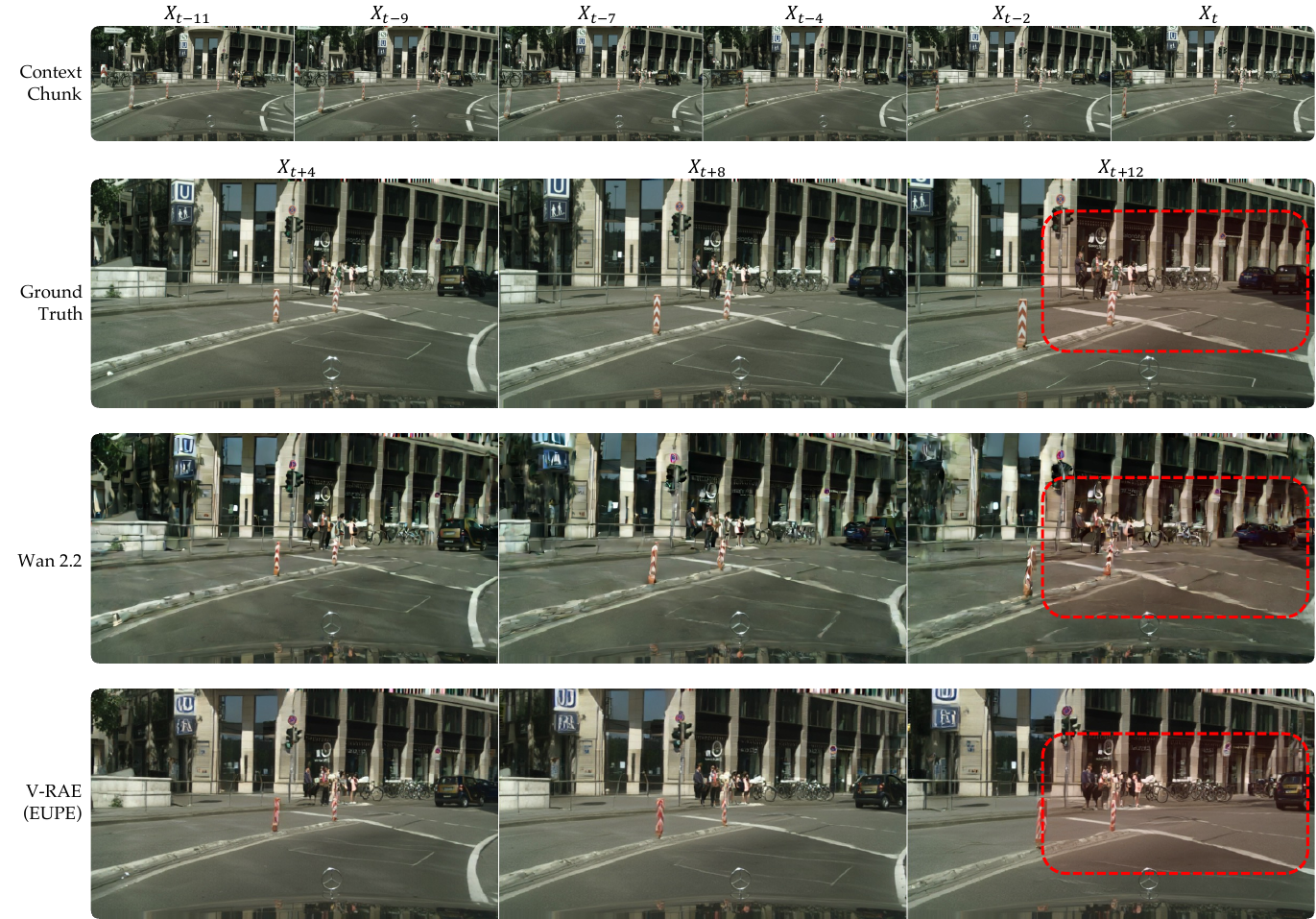}
    \par
    {\small (b) Pedestrian-dominated intersection scene}\par
    \caption{\textbf{Qualitative comparison of future video prediction on Cityscapes.} We show (a) a traffic scene dominated by vehicle motion and (b) an intersection scene containing pedestrians and bicycles. In each panel, the top row contains six context frames, followed by ground-truth future frames, predictions using the Wan2.2 VAE baseline, and V-RAE predictions. The three future frames correspond to successive latent chunks at $X_{t+4}$, $X_{t+8}$, and $X_{t+12}$. Red dashed boxes highlight dynamic regions where the two latent spaces differ most clearly.}
    \label{fig:world-modeling}
\end{figure}

\vspace{-2mm}
\paragraph{Experimental Setup.}
We evaluate on the Cityscapes driving dataset~\citep{cordts2016cityscapes}, conditioning on frames 4-15 and predicting frames 16-27 at a resolution of $432\times768$.
V-RAE leverages EUPE-B~\citep{zhu2026eupe} as its representation encoder and is also fine-tuned on CoVLA~\citep{arai2025covla} for high-resolution video reconstruction.
We take the Wan2.2 VAE~\citep{wan2025wan} as the conventional latent baseline.
Both latent spaces share the same Transformer backbone, latent-token budget, training schedule, and sampling configuration, with only the latent-channel-dependent input and output projections adjusted.
We compute gFID~\citep{heusel2017gans} over 6,000 future frames generated from 500 validation sequences, and gFVD~\citep{unterthiner2018fvd} over the corresponding 500 12-frame videos.

\begin{table}[!t]
    \centering
    \caption{\textbf{Future video prediction results on Cityscapes at 40k training steps.} Both methods use the same conditional DiT and training budget.}
    \label{tab:world-modeling}
    \setlength{\tabcolsep}{8pt}
    \begin{tabular}{lcccc}
        \toprule
        \textbf{Latent space} & \textbf{rFVD$\downarrow$} & \textbf{tFVD$\downarrow$} & \textbf{gFID$\downarrow$} & \textbf{gFVD$\downarrow$} \\
        \midrule
        Wan2.2 VAE & \textbf{7.0256} & 319.0233 & 15.02 & 144.47 \\
        V-RAE (EUPE-B) & 29.2931 & \textbf{224.6040} & \textbf{11.52} & \textbf{111.36} \\
        \bottomrule
    \end{tabular}
\end{table}

\vspace{-2mm}
\paragraph{Results and Analysis.}
Under the same conditional DiT and training budget, V-RAE reduces gFID from \textbf{15.02} to \textbf{11.52} and gFVD from \textbf{144.47} to \textbf{111.36}, demonstrating that its latent space is more effective for modeling future video distributions.
Despite its worse reconstruction fidelity (29.2931 vs.\ 7.0256 rFVD), V-RAE achieves a substantially lower tFVD (224.6040 vs.\ 319.0233) together with superior future prediction.
This reversal echoes Section~\ref{sec:reconstruction-vs-generation}: tFVD probes temporal smoothness and decoder robustness to off-trajectory latent states, properties that are more relevant than reconstruction fidelity when a predictor must generate approximate future codes.
Because the downstream architecture and optimization budget are controlled, the results support attributing much of the performance gap to the choice of latent representation.
Figure~\ref{fig:world-modeling} provides consistent qualitative evidence: both methods preserve static road and building layouts, but the Wan2.2 baseline increasingly blurs vehicles, pedestrians, and bicycles and distorts their relative positions at longer horizons, whereas V-RAE better maintains object contours, lane structure, and instance separation.
The lower tFVD and improved qualitative stability indicate that a semantically organized and temporally smooth latent space reduces structural and identity drift during future-state prediction, leading to more realistic and coherent videos.

\finding{4}{Semantic latents form a directly decodable predictive state space.}{
A predictor learns transitions between semantic states, while the same frozen
decoder renders predicted states into pixels, unifying future-state prediction
and visual reconstruction within a single latent interface.}

\section{Related Work}

\paragraph{Representation Learning for Generation.}
Pretrained visual representations have increasingly been incorporated into generative models.
In the image domain, prior work aligns tokenizer latents with pretrained features~\citep{ma2025unitok,wu2025towards}, aligns generator features through auxiliary objectives~\citep{yu2024repae,leng2025repae,wu2025representation}, or directly adopts frozen encoder features as the generative latent space~\citep{zheng2025rae,singh2026raev2,tong2026scaling_rae}.
In video, DERA~\citep{guo2025dera} aligns tokenizer representations, while VideoREPA~\citep{zhang2025videorepa} and \citet{wu2025representation} align diffusion features.
Divot~\citep{ge2025divot} models continuous video-tokenizer features for unified understanding and generation, but learns those features specifically as a video tokenizer rather than inheriting a frozen VFM feature space.
To the best of our knowledge, V-RAE is the first to perform video generative modeling directly in a temporally compressed, high-dimensional semantic feature space inherited from frozen vision foundation models, with generated latents decoded into pixel-space videos.

\vspace{-2mm}
\paragraph{Video Autoencoders.}
Video autoencoders provide the latent spaces used by modern video generators~\citep{blattmann2023stable}.
Representative designs include continuous video VAEs~\citep{wan2025wan,kong2024hunyuanvideo,yang2025cogvideox,wang2024vidtwin}, discrete tokenizers~\citep{yu2024magvitv2,wang2024omnitokenizer,kondratyuk2023videopoet,agarwal2025cosmos}, and query-based tokenizers~\citep{wang2025larp,tan2025sweettok,xiong2026evatok}.
These approaches primarily optimize compact representations exclusively for pixel-level reconstruction, leaving their latent spaces devoid of semantic structure. 
In contrast, V-RAE starts from a frozen semantic representation space and learns temporal compression and pixel decoding, while preserving sufficient semantics and producing smooth temporal trajectories.

\section{Conclusion}
We introduce V-RAE, a video representation autoencoder that repurposes frozen visual representation models as semantically structured latent spaces for video reconstruction and generation.
V-RAE achieves near-best reconstruction on UCF101 and outperforms all evaluated video autoencoders on K600, while retaining substantially richer semantic information than conventional VAE-based video tokenizers.
When used as the latent space of a DiT, V-RAE further improves generation quality and converges up to \(6\times\) faster under matched training settings.
Our analysis also reveals that reconstruction and generation metrics induce markedly different rankings of video autoencoders, indicating that pixel reconstruction quality alone is insufficient to characterize the suitability of a latent space for generation.
To better diagnose this property, we introduced tFVD, which evaluates decoded trajectories between latent samples and more reliably reflects downstream generation performance than rFVD.
Preserving the semantic and temporal organization of pretrained representations therefore provides an effective alternative to learning video latents solely through reconstruction objectives.

{
    \small
    \bibliographystyle{plainnat}
    \bibliography{main}
}

\newpage
\appendix
\raggedbottom

\section{Implementation Details}
\label{app:implementation}

We provide detailed model and training configurations for reproducibility.
Section~\ref{app:vrae-architecture} describes the V-RAE architecture and temporal pooling designs, Section~\ref{app:reconstruction-training} specifies the reconstruction recipe and shared VideoMAE discriminator, and
Section~\ref{app:latent-dit} reports the generation architecture, training, sampling, and evaluation settings.

\subsection{V-RAE Architecture}
\label{app:vrae-architecture}

\begin{table}[H]
  \centering
  \captionsetup{skip=5pt}
  \caption{VFM-dependent V-RAE architecture. Block indices are zero-based;
  shared pooler and decoder details are described in text.}
  \label{tab:app-vrae}
  \scriptsize
  \setlength{\tabcolsep}{3.2pt}
  \renewcommand{\arraystretch}{1.12}
  \begin{tabularx}{\linewidth}{@{}p{0.245\linewidth}YYYY@{}}
    \toprule
    {\footnotesize\textbf{Configuration}} & {\footnotesize\textbf{DINOv3-L}} & {\footnotesize\textbf{SigLIP2-L}} & {\footnotesize\textbf{EUPE-B}} & {\footnotesize\textbf{V-JEPA~2.1-L}} \\
    \midrule
    \grouprow{5}{{\footnotesize Encoder settings}}
    \addlinespace[2.5pt]
    Backbone (blocks / width) & ViT-L/16 (24 / 1024) & ViT-L/16-256 (24 / 1024) & ViT-B/16 (12 / 768) & ViT-L/16 (24 / 1024) \\
    Selected feature blocks & 11,13,15,17,19,21,23 & 11,13,15,17,19,21,23 & 5,6,7,8,9,10,11 & 11,13,15,17,19,21,23 \\
    Patch / encoder tubelet & 16 / 1 & 16 / 1 & 16 / 1 & 16 / 2 \\
    Input norm. & ImageNet-style & Native processor & ImageNet-style & ImageNet-style \\
    Encoder temporal length & $T$ & $T$ & $T$ & $T/2$ \\
    \midrule
    \grouprow{5}{{\footnotesize Decoder settings}}
    \addlinespace[2.5pt]
    Temporal group size & 4 & 4 & 4 & 2 \\
    Latent width $C$ & 1024 & 1024 & 768 & 1024 \\
    Decoder initialization & RAEv2 image decoder & Scratch & RAEv2 image decoder & Scratch \\
    Decoder attention mask & Chunk-causal & Chunk-causal & Chunk-causal & Full (non-causal) \\
    Trainable Params & \multicolumn{4}{c}{$\sim$420M} \\
    \bottomrule
  \end{tabularx}
\end{table}

\paragraph{Representation encoders.}
We instantiate V-RAE with DINOv3-L, SigLIP2-L, EUPE-B, and V-JEPA~2.1-L and keep the representation encoder frozen throughout reconstruction training.
DINOv3-L, SigLIP2-L, and V-JEPA~2.1-L use 1024-dimensional features, whereas EUPE-B uses 768-dimensional features.
The three image encoders preserve the input temporal length, while the tubelet size of two in V-JEPA~2.1-L reduces it by half.
We therefore pool groups of four image-encoder features or two V-JEPA features, producing the same temporal compression ratio of four for all variants.
A 16-frame, $256\times256$ clip is represented by a $4\times16\times16$ latent grid with channel width $C$ given in Table~\ref{tab:app-vrae}.

\paragraph{Non-affine latent normalization in temporal pooling.}
The temporal attention operation is described in the main text; here we focus on the normalization applied to its output.
Let $\tilde{\bm{z}}$ denote the pooled feature. Before latent-noise augmentation, we apply LayerNorm without a learned scale or bias:
\begin{equation}
  \bm{z}_{\mathrm{clean}}=\operatorname{LN}_{\mathrm{na}}(\tilde{\bm{z}}),\qquad
  \bm{z}_{\mathrm{train}}=\bm{z}_{\mathrm{clean}}+\sigma\epsilon.
\end{equation}
This choice is important for noise-augmented reconstruction training.
With an affine LayerNorm, $\bm{z}_{\mathrm{clean}}=\gamma\odot\operatorname{LN}(\tilde{\bm{z}})+\beta$, the optimizer can increase $\gamma$ while the injected noise magnitude $\sigma$ remains fixed.
The effective signal-to-noise ratio then grows approximately as $\gamma^2/\sigma^2$, providing a shortcut that reduces the relative corruption
without requiring the decoder to become robust to noise.
We therefore fix $\gamma=1$ and $\beta=0$.
This removes the rescaling shortcut, keeps the latent scale calibrated, and forces the decoder to reconstruct under the intended
noise level.
With identity-initialized projections and zero-initialized queries and temporal biases, the pooler initially reduces to temporal mean pooling followed by this fixed normalization.

\paragraph{V-RAE decoder.}
All variants share a ViT-XL video decoder with width 1152, 28 transformer blocks, 16 attention heads, and a 4096-dimensional GELU feed-forward network.
The decoder combines a learned 2D spatial embedding with 3D RoPE ($\theta=10{,}000$).
The DINOv3-L, SigLIP2-L, and EUPE-B reconstruction models use chunk-causal attention: tokens interact freely within a latent chunk, while each chunk can attend only to itself and earlier chunks.
The V-JEPA~2.1-L model instead uses non-causal full self-attention over all spatiotemporal latent tokens, consistent with the full temporal context provided by its frozen encoder.
Each output token predicts a $4\times16\times16\times3=3072$-dimensional RGB tubelet, which is rearranged into four consecutive frames.
For DINOv3-L and EUPE-B, compatible transformer weights are initialized from the corresponding RAEv2 image decoder, and the image prediction head is inflated along time; the other two decoders are trained from scratch.

\paragraph{Temporal pooling designs.}
\label{app:pooling-designs}

\begin{figure}[!t]
  \vspace{-12pt}
  \centering
  \begin{minipage}[t]{0.24\linewidth}
    \centering
    \parbox[b][45mm][b]{\linewidth}{%
      \centering
      \includegraphics[width=0.73\linewidth]{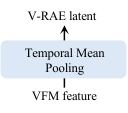}%
    }
    \par\vspace{2pt}
    {\scriptsize\strut (a) Mean pooling\strut\par}
  \end{minipage}\hfill%
  \begin{minipage}[t]{0.24\linewidth}
    \centering
    \parbox[b][45mm][b]{\linewidth}{%
      \centering
      \includegraphics[width=0.73\linewidth,trim=3.5bp 0.20bp 3.5bp 0,clip]{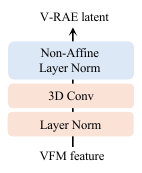}%
    }
    \par\vspace{2pt}
    {\scriptsize\strut (b) 3D convolutional pooling\strut\par}
  \end{minipage}\hfill%
  \begin{minipage}[t]{0.24\linewidth}
    \centering
    \parbox[b][45mm][b]{\linewidth}{%
      \centering
      \includegraphics[width=0.79\linewidth,trim=10.5bp 0.20bp 12.5bp 0,clip]{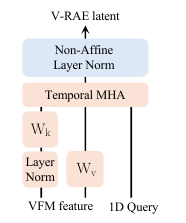}%
    }
    \par\vspace{2pt}
    {\scriptsize\strut (c) Temporal attention pooling\strut\par}
  \end{minipage}\hfill%
  \begin{minipage}[t]{0.24\linewidth}
    \centering
    \parbox[b][45mm][b]{\linewidth}{%
      \centering
      \includegraphics[width=0.74\linewidth,trim=8.5bp 0.13bp 10.5bp 0,clip]{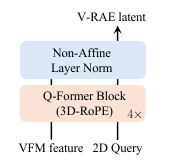}%
    }
    \par\vspace{2pt}
    {\scriptsize\strut (d) Q-Former pooling\strut\par}
  \end{minipage}
  \caption{Structures of the four temporal pooling designs. All modules reduce the encoder features to the same temporal latent length, differing only in how features within each temporal group are aggregated.}
  \label{fig:app-pooling-designs}
  \vspace{-2mm}
\end{figure}

Temporal compression must remove frame-level redundancy without destroying the semantic geometry inherited from the frozen representation encoder.
We study this trade-off along two design axes: whether aggregation is content-adaptive, and how strongly the pooler transforms the encoder feature space.
For a fair comparison, we fix the DINOv3-L encoder, video decoder, training data, temporal compression ratio, and reconstruction objective.
Every module maps one non-overlapping temporal group to one latent feature map on the same spatial grid; only the aggregation operator changes. Figure~\ref{fig:app-pooling-designs}
summarizes the resulting progression from fixed averaging to learned latent resampling.

The four designs expose different inductive biases.
Mean pooling in Figure~\ref{fig:app-pooling-designs}(a) is parameter-free and leaves the feature basis untouched, making it a useful reference for semantic preservation.
However, it assigns the same weight to every frame and cannot distinguish a stable region from rapid motion or changes in appearance.
The 3D convolution in Figure~\ref{fig:app-pooling-designs}(b) instead learns a shared local spatiotemporal filter.
This favors the reconstruction of local details, but its aggregation weights remain fixed after training, and convolution can modify the spatial organization of the pretrained features.

Temporal attention in Figure~\ref{fig:app-pooling-designs}(c) retains the same spatial aggregation constraint while making the temporal weights depend on the input content.
A single 1D query per head is shared across spatial positions and temporal groups, so the module learns which frames to retain without introducing a separate query grid or a deep feature transformation.
Identity-initialized projections and zero-initialized queries and temporal biases make its initial behavior close to mean pooling; optimization can then depart from this conservative starting point only when reconstruction provides evidence for doing so.
The final non-affine LayerNorm fixes the output scale for the noise-augmented decoder training described above.

Q-Former pooling in Figure~\ref{fig:app-pooling-designs}(d) represents the opposite end of the capacity spectrum. Its learned 2D query grid is refined by four transformer blocks with 3D RoPE, allowing repeated spatial and temporal feature rewriting before producing the latent.
This flexibility is useful when strong latent resampling is required, but is unnecessarily heavy for pooling short temporal groups and can move the output farther from the frozen encoder's semantic feature space.
The ablation in Figure~\ref{fig:temporal-pooling-ablation} reflects these trade-offs: mean pooling best preserves probing accuracy but reconstructs poorly, convolutional pooling improves reconstruction at a substantial semantic cost, and the much larger Q-Former does not improve the overall balance.
Temporal attention offers content-adaptive aggregation with only about 3M parameters and provides the best practical compromise, so we use it throughout the remaining experiments.

\subsection{Reconstruction Training Recipe}
\label{app:reconstruction-training}

\paragraph{Reconstruction training.}
We jointly train the temporal pooler and video decoder while keeping the representation encoder frozen in evaluation mode.
Training uses the union of UCF101 train split~1 and the K600 training split.
We sample 16-frame clips at an interval of 3 from a random temporal position, resize the short side to 256, and apply a random $256\times256$ crop and horizontal flip.
DINOv3-L and EUPE-B are trained for 40 epochs, while SigLIP2-L and V-JEPA~2.1-L are trained for 50 epochs.
The remaining shared training hyperparameters are listed in
Table~\ref{tab:app-recon-training}.

\begin{table}[H]
  \centering
  \caption{Shared reconstruction training hyperparameters.}
  \label{tab:app-recon-training}
  \footnotesize
  \setlength{\tabcolsep}{6pt}
  \renewcommand{\arraystretch}{1.08}
  \begin{tabularx}{0.78\linewidth}{@{}>{\centering\arraybackslash}p{0.31\linewidth}>{\centering\arraybackslash}X@{}}
    \toprule
    \textbf{Configuration} & \textbf{Value} \\
    \midrule
    Global batch size & 128 \\
    Optimizer & AdamW \\
    Learning rate & $2\times10^{-4}$ \\
    Betas & $(0.9,0.95)$ \\
    Epsilon & $10^{-8}$ \\
    Weight decay & 0 \\
    Learning-rate schedule & Constant \\
    Gradient clipping & Global norm 1.0 \\
    Precision & BF16 \\
    EMA decay & 0.9995 \\
    L1 loss weight & 1 \\
    LPIPS loss weight & 1 \\
    GAN generator loss weight & 0.3 \\
    Gram loss weight & 100 \\
    \bottomrule
  \end{tabularx}
\end{table}

\paragraph{Reconstruction objective.}
For each clip, the frozen encoder and temporal pooler produce a clean latent $\bm{z}_{\mathrm{clean}}$.
We perturb it during reconstruction training as
$\bm{z}_{\mathrm{train}}=\bm{z}_{\mathrm{clean}}+\sigma\epsilon$, where
$\sigma\sim\mathcal{U}(0,0.8)$ and $\epsilon\sim\mathcal{N}(0,I)$.
The video decoder is optimized with L1, LPIPS, GAN, and Gram losses, whose weights are listed in Table~\ref{tab:app-recon-training}.
LPIPS is evaluated at the native $256\times256$ resolution, whereas the Gram branch uses $224\times224$ inputs; both perceptual branches cover all 16 frames.
The latent perturbation is used only as a reconstruction-training augmentation; evaluation, latent-statistics estimation, and latent video generation all use clean latents.

\begin{table}[H]
  \centering
  \caption{Shared VideoMAE discriminator configuration.}
  \label{tab:app-disc}
  \footnotesize
  \setlength{\tabcolsep}{6pt}
  \renewcommand{\arraystretch}{1.08}
  \begin{tabularx}{0.78\linewidth}{@{}>{\centering\arraybackslash}p{0.31\linewidth}>{\centering\arraybackslash}X@{}}
    \toprule
    \textbf{Configuration} & \textbf{Value} \\
    \midrule
    Backbone & Frozen VideoMAE-B \\
    Feature taps & Embedding output and blocks 2, 5, 8, 11 \\
    Temporal head & Conv1d, kernel size 9 \\
    Normalization & Local batch normalization + spectral normalization \\
    Adversarial start & Epoch 30 \\
    Update interval & One discriminator step per five generator steps \\
    Optimizer & AdamW \\
    Learning rate & $2\times10^{-5}$ \\
    Learning-rate schedule & Constant \\
    Betas & $(0.9,0.95)$ \\
    Weight decay & $10^{-4}$ \\
    \bottomrule
  \end{tabularx}
\end{table}

\paragraph{Adversarial objective.}
Beginning at epoch 30, we add the same VideoMAE-based adversarial objective for all four variants.
The VideoMAE-B backbone remains frozen, and only the temporal Conv1d heads are optimized.
We use a non-saturating generator loss and a smoothed non-saturating discriminator loss, both weighted by 0.3, and update the discriminator once every five generator steps.
Translation, color, and cutout DiffAug are applied with probability 1.0 and cutout ratio 0.2.
We additionally use LeCam regularization with weight 0.001 and logit-EMA decay 0.9995.

\subsection{Generation Training Recipe}
\label{app:latent-dit}

\paragraph{DiT architecture.}
Our goal is to compare the latent spaces induced by different autoencoders, rather than the capacity of their downstream generators.
We therefore use a high-capacity DiT to reduce the risk that limited generator expressiveness becomes the dominant bottleneck.
The backbone and latent-token budget are matched across autoencoders, so differences in generation quality and convergence more directly reflect how readily each latent distribution can be modeled.
Table~\ref{tab:app-dit} summarizes the shared latent DiT architecture and the task-specific configurations for UCF101, K600, and Cityscapes.

\begin{table}[H]
  \centering
  \captionsetup{skip=5pt}
  \caption{Latent video DiT configurations for UCF101, K600, and Cityscapes.}
  \label{tab:app-dit}
  \scriptsize
  \setlength{\tabcolsep}{3.2pt}
  \renewcommand{\arraystretch}{1.12}
  \begin{tabularx}{\linewidth}{@{}p{0.29\linewidth}YYY@{}}
    \toprule
    {\footnotesize\textbf{Configuration}} & {\footnotesize\textbf{UCF101 generation}} & {\footnotesize\textbf{K600 generation}} & {\footnotesize\textbf{Cityscapes prediction}} \\
    \midrule
    \grouprow{4}{{\footnotesize Input and token geometry}}
    \addlinespace[2.5pt]
    RGB input & $20\times256^2$, interval 3 & $20\times256^2$, interval 3 & Frames 4--15 $\to$ 16--27, $432\times768$ \\
    Clean latent grid & $5\times16\times16\times C$ & $5\times16\times16\times C$ & $3\times27\times48\times768$ each \\
    Encoder / output tokens & 1280 / 1280 & 1280 / 1280 & 7776 / 3888 \\
    Latent patch size & $1\times1$ & $1\times1$ & $1\times1$ \\
    \midrule
    \grouprow{4}{{\footnotesize DiT encoder and output decoder}}
    \addlinespace[2.5pt]
    Main encoder width / blocks & 1536 / 28 & 1536 / 28 & 1152 / 28 \\
    Main encoder heads / head dim & 24 / 64 & 24 / 64 & 16 / 72 \\
    Output decoder width / blocks & 2048 / 2 & 2048 / 2 & 2048 / 2 \\
    Output decoder heads / head dim & 16 / 128 & 16 / 128 & 16 / 128 \\
    Self-attention & Full spatiotemporal & Full spatiotemporal & Full spatiotemporal \\
    Positional encoding & 3D RoPE ($\theta=10{,}000$) & 3D RoPE, ($\theta=10{,}000$) & 3D RoPE, ($\theta=10{,}000$) \\
    Block conditioning & AdaLN-Zero & AdaLN-Zero & AdaLN-Zero \\
    Normalization / epsilon & RMSNorm / $10^{-6}$ & RMSNorm / $10^{-6}$ & RMSNorm / $10^{-6}$ \\
    FFN / MLP ratio & SwiGLU / 4.0 & SwiGLU / 4.0 & SwiGLU / 4.0 \\
    Time embedding & 256-D Gaussian Fourier & 256-D Gaussian Fourier & 256-D Gaussian Fourier \\
    Auxiliary base branch & Encoder block 8 & Encoder block 8 & Encoder block 8 \\
    Trainable parameters & 1.368B & 1.369B & 0.843B \\
    \midrule
    \grouprow{4}{{\footnotesize Conditioning}}
    \addlinespace[2.5pt]
    Condition & Class label & Class label & Clean context latent \\
    Number of classes & 101 & 600 & \na \\
    Condition dropout & 0.1 & 0.1 & 0.1 \\
    Context embedding & \na & \na & Dropout $\rightarrow$ mean pool $\rightarrow$ 2-layer MLP \\
    \midrule
    \grouprow{4}{{\footnotesize Optimization}}
    \addlinespace[2.5pt]
    Global batch size & 64 & 128 & 8 \\
    Optimizer / learning rate & AdamW / $10^{-4}$ & AdamW / $10^{-4}$ & AdamW / $3\times10^{-4}$ \\
    Weight decay & 0 & 0 & 0 \\
    LR schedule & Linear to $5\times10^{-5}$ & Linear to $5\times10^{-5}$ & Cosine to $10^{-4}$ \\
    Warmup & 10 epochs from zero & 2 epochs from zero & 1,490 updates from zero \\
    Gradient clipping & Global norm 1.0 & Global norm 1.0 & Global norm 1.0 \\
    Precision & BF16 & BF16 & BF16 \\
    Training length / EMA & 1700 epochs / 0.9995 & 100 epochs / 0.9995 & 40k steps / 0.9995 \\
    \midrule
    \grouprow{4}{{\footnotesize Flow matching}}
    \addlinespace[2.5pt]
    Prediction parameterization & $x$-prediction & $x$-prediction & $x$-prediction \\
    Time distribution & LogitNormal$(0,1)$ & LogitNormal$(0,1)$ & LogitNormal$(0,1)$ \\
    Velocity denominator clamp $t_\epsilon$ & 0.05 & 0.05 & 0.05 \\
    Full / base loss weights & 1.0 / 1.0 & 1.0 / 1.0 & 1.0 / 1.0 \\
    \midrule
    \grouprow{4}{{\footnotesize Online sampling and evaluation}}
    \addlinespace[2.5pt]
    Online sampler / steps & Euler / 100 & Euler / 100 & Euler / 100 \\
    Online CFG / internal guidance & 1.0 / 1.3 & 1.0 / 1.2 & 1.0 / 1.2 \\
    Evaluation population & 2048 clips & 50,000 clips & 500 clips \\
    Video metric & 17-frame gFVD & 17-frame gFVD & 12-frame gFID + gFVD \\
    \bottomrule
  \end{tabularx}
\end{table}

For UCF101 and K600, all four representation encoders use the same class-conditional DiT backbone: a 28-block, 1536-dimensional transformer encoder and a two-block, 2048-dimensional output decoder.
Only the input channel count and corresponding time shift change for EUPE-B.
Cityscapes retains the 28-block depth but uses a 1152-dimensional encoder with 16 attention heads of dimension 72 and EUPE-B latents with 768 channels.
It uses a clean-context/noisy-future token sequence; the output decoder predicts only the future tokens.
All variants use full self-attention, per-head Q/K RMSNorm, 3D RoPE, AdaLN-Zero conditioning,
RMSNorm, and SwiGLU.

For conditioning, UCF101 and K600 use class embeddings with classifier-free dropout probability 0.1. Cityscapes instead conditions on clean context latents; the context is randomly dropped with probability 0.1, mean pooled,
and mapped by a two-layer MLP. A secondary prediction branch is attached after
encoder block 8 and is used for the auxiliary training objective and internal
guidance.

\paragraph{Training details.}
We freeze the complete V-RAE and compute latent statistics separately for each dataset and encoder. Cityscapes uses the CoVLA-fine-tuned EUPE-B V-RAE checkpoint and statistics computed from training future latents only.
UCF101 trains for 1,700 epochs with batch size 64 and a 10-epoch warmup; K600 trains for 100 epochs with batch size 128 and a 2-epoch warmup. Both linearly decay the learning rate from $10^{-4}$ to $5\times10^{-5}$.
Cityscapes instead runs for 40,000 steps with batch size 8, a 1,490-update linear warmup to $3\times10^{-4}$, and cosine decay to $10^{-4}$.
All configurations use BF16 mixed precision and an EMA decay of 0.9995.
For flow matching, given a normalized clean latent $\bm{x}_0$ and Gaussian noise $\bm{\epsilon}$, we form the interpolated latent
\begin{equation}
  \bm{x}_t=(1-t) \cdot \bm{x}_0+t \cdot \epsilon\;\;,\qquad
  t=\frac{s \cdot t_0}{1+(s-1) \cdot t_0}\;\;,\qquad
  t_0\sim\operatorname{LogitNormal}(0,1)\;.
\end{equation}
The time shift $s$ is calculated from the latent dimensionality as defined in Equation~(\ref{eq:time-shift}).
The DiT predicts $\bm{x}_0$ and converts the prediction to velocity before evaluating the flow-matching MSE.
The final objective sums the main prediction loss and the auxiliary loss from encoder block~8 with unit weight for each term.

\paragraph{Sampling and evaluation.}
We sample all generation and prediction models from their EMA weights using a 100-step Euler solver from $t=1$ to $t=0$. Online samples use CFG 1.0, with internal guidance 1.3 for UCF101 and 1.2 for K600 and Cityscapes active over $t\in[0.10,1.0]$.
Reported results are computed separately from online monitoring, using the benchmark-specific frame ranges and evaluation populations described below.

\section{Evaluation Protocols}
\label{app:evaluation}

Unless stated otherwise, we evaluate EMA checkpoints.
The rFVD, tFVD, and gFVD metrics use clip features from an I3D network pretrained on Kinetics-400, whereas gFID uses Inception-v3 pool3 features.
All Fr\'echet means and covariances are accumulated in float64.

\paragraph{Reconstruction.}
We evaluate on the UCF101 test set and the K600 validation set.
Each clip is taken from the beginning of a video with a temporal interval of 3; we resize the short side and apply a center crop to $256\times256$.
V-RAE and AToken use
16-frame clips, while causal VAEs that encode the first frame separately use 17-frame clips, following Table~\ref{tab:reconstruction}.
K600 contributes 27,874 clips after removing 36 videos that are too short; no temporal padding is used. LPIPS, PSNR, and SSIM are evaluated over all valid frames, and rFVD is computed between the reconstructed and real clip distributions.

\paragraph{Temporal interpolation.}
For tFVD, we construct a temporally ordered sequence of six clean latent codes
$[\bm{z}_0,\ldots,\bm{z}_5]$ using each tokenizer's native temporal grouping.
Here, $\bm{z}_0$ is the separately encoded first-frame latent for causal VAEs and the latent chunk of the first four frames for V-RAE; later codes follow the tokenizer's native temporal order.
We replace the four interior codes with local midpoints:
\begin{equation}
  \widetilde{\bm{Z}}=\left[
  \frac{\bm{z}_0+\bm{z}_2}{2},
  \frac{\bm{z}_1+\bm{z}_3}{2},
  \frac{\bm{z}_2+\bm{z}_4}{2},
  \frac{\bm{z}_3+\bm{z}_5}{2}\right].
\end{equation}
The four interpolated codes are decoded from temporal position zero, producing a 16-frame clip.
We compare it with the aligned ground-truth frames 4-19 of the input clip and compute tFVD using the same I3D implementation as rFVD.

\paragraph{Class-conditional generation.}
All tokenizer comparisons are evaluated on 17-frame videos at $256\times256$.
V-RAE and AToken first generate 20 frames and retain the first 17 for metric computation; causal VAEs with a separate first-frame encoder generate 17 frames directly.
UCF101 gFVD uses 2,048 generated clips, while K600 uses 50,000. Within each benchmark, every tokenizer uses the same evaluation population and the same 100-step Euler sampling protocol.
V-RAE samples use EMA DiT weights and CFG 1.0; the internal-guidance scale is fixed within each
reported comparison.

\paragraph{Future prediction.}
For Cityscapes, the model conditions on frames 4-15 and predicts frames 16-27 at $432\times768$ and 16 fps.
We evaluate all 500 validation sequences.
The same 12-frame predictions are used for both metrics: gFID is computed over the resulting 6,000 frames, and gFVD over the 500 video clips.
Sampling uses the EMA model, 100 Euler steps, CFG 1.0, and internal guidance 1.2 over $t\in[0.10,1.0]$.

\section{Additional Qualitative Results}
\label{app:additional-generation-results}

\begin{figure}[H]
  \centering
  \includegraphics[width=\linewidth]{\detokenize{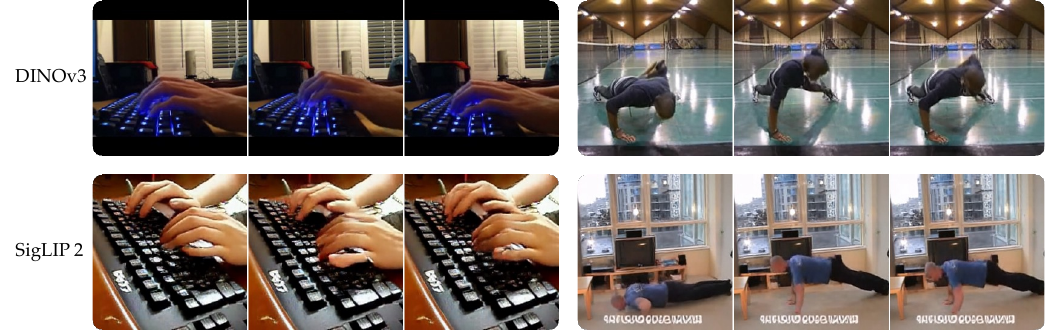}}
  \vspace{-4pt}
  {\small (a) UCF101}
  \vspace{5pt}

  \includegraphics[width=\linewidth]{\detokenize{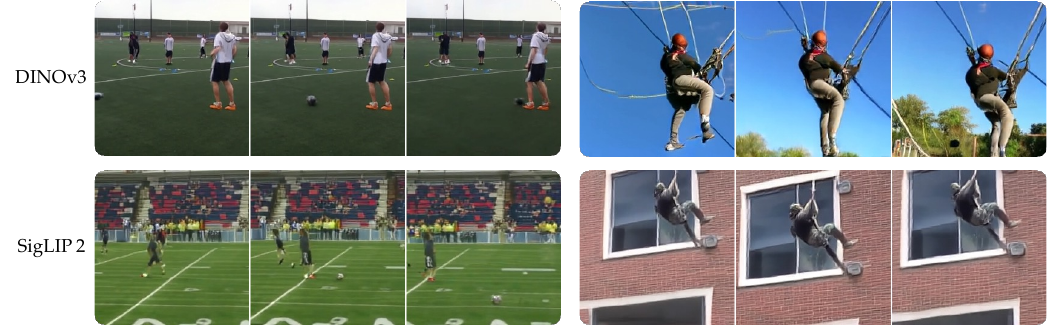}}
  \vspace{-4pt}
  {\small (b) Kinetics-600}
  \caption{Class-conditional generation results for V-RAE with
  DINOv3-L and SigLIP2-L}
  \label{fig:app-generation-dino-siglip}
\end{figure}

Figure~\ref{fig:app-generation-dino-siglip} provides additional samples from the DINOv3-L and SigLIP2-L variants of V-RAE.
Together with Figures~\ref{fig:generation-visualization-ucf101} and~\ref{fig:generation-visualization-k600}, these results cover all four V-RAE representation encoders evaluated in our generation experiments.

\section{Limitations}
\label{app:limitations}

We finally discuss several limitations of the current work that motivate future research.
First, our experiments primarily focus on relatively controlled video benchmarks and moderate-scale generative models.
While these settings allow us to isolate the effect of the latent representation, it remains unclear whether the advantages of V-RAE extend to large-scale open-domain text-to-video generation with larger datasets, higher resolutions, longer videos, and substantially larger DiT backbones.
Second, although we evaluate several representative image- and video-based encoders, we do not systematically disentangle how different pretraining objectives, architectures, and spatiotemporal inductive biases affect reconstruction and generation.
Finally, while our experiments reveal a strong association between semantic preservation and downstream generation quality, the precise relationship between semantic structure in the latent space and its generative utility remains to be better understood.

\end{document}